\pdfoutput=1
\PassOptionsToPackage{unicode}{hyperref}
\PassOptionsToPackage{hyphens}{url}
\PassOptionsToPackage{dvipsnames,svgnames,x11names}{xcolor}
\documentclass[
  12pt]{article}

\usepackage{amsmath,amssymb}
\usepackage{mathtools,amsthm}
\usepackage[T1]{fontenc}
\usepackage[utf8]{inputenc}
\usepackage{textcomp}
\usepackage{lmodern}
\IfFileExists{upquote.sty}{\usepackage{upquote}}{}
\IfFileExists{microtype.sty}{%
  \usepackage[]{microtype}
  \UseMicrotypeSet[protrusion]{basicmath} %
}{}
\makeatletter
\@ifundefined{KOMAClassName}{%
  \setlength{\parindent}{2em}
  \setlength{\parskip}{0pt}
}{%
  \KOMAoptions{parskip=false}
  \setlength{\parindent}{2em}}
\makeatother
\usepackage{indentfirst}
\usepackage{xcolor}
\usepackage{booktabs,longtable,array}
\usepackage{graphicx}
\usepackage{caption}
\usepackage{placeins,needspace,flafter}
\usepackage{algorithm}
\makeatletter\renewcommand\floatc@ruled[2]{{\normalcolor\@fs@cfont #1} {\normalcolor #2}\par}\makeatother %
\usepackage[noend]{algpseudocode}
\usepackage{enumitem,etoolbox}
\usepackage[authoryear,round]{natbib}
\usepackage{hyperref}
\usepackage{bookmark}
\IfFileExists{xurl.sty}{\usepackage{xurl}}{}
\def\spacingset#1{\renewcommand{\baselinestretch}{#1}\small\normalsize}

\AtBeginEnvironment{tabular}{\renewcommand{\baselinestretch}{1}\selectfont}
\AtBeginEnvironment{longtable}{\renewcommand{\baselinestretch}{1}\selectfont}
\AtBeginEnvironment{algorithm}{\spacingset{1}}

\newtheorem{proposition}{Proposition}
\newtheorem{theorem}{Theorem}

\newtheorem{remark}{Remark}

\newcommand{\E}{\mathbb{E}}
\newcommand{\Var}{\mathrm{Var}}

\newcommand{\Gid}{G_{\mathrm{id}}}
\newcommand{\rN}{r_{\mathrm{N}}}
\newcommand{\rhoN}{\rho_{\mathrm N}}
\newcommand{\kapN}{\kappa_{\mathrm N}}
\newcommand{\ind}{\mathbf 1}
\newcommand{\cN}{\mathcal N}

\def\anon{1}
\date{}
\hypersetup{
  pdftitle={Optimal Labelling Designs for Sequential Decisions on Language Model Updates},
  pdfkeywords={Active sampling; e-values; LLM judge; Neyman allocation; Prediction-powered inference; Testing by betting},
  colorlinks=true,
  linkcolor=blue,
  filecolor=Maroon,
  citecolor=Blue,
  urlcolor=Blue
}
\if1\anon
  \hypersetup{pdfauthor={Doudou Zhou, Jiahui Xie, Yue Liu}}
\else
  \hypersetup{pdfauthor={}}
\fi

\begin{document}
\spacingset{1}

\if1\anon
\title{\bfseries Optimal Labelling Designs for Sequential Decisions on Language Model Updates}
\author{%
Doudou Zhou$^{1}$\thanks{Corresponding author. Email: ddzhou@nus.edu.sg.},
Jiahui Xie$^{1}$ and
Yue Liu$^{2}$\\[0.35em]
{\small $^{1}$Department of Statistics and Data Science, National University of Singapore}\\
{\small $^{2}$Department of Statistics, Harvard University}%
}
\maketitle
\fi

\if0\anon
\bigskip\bigskip\bigskip
\begin{center}
{\LARGE\bfseries Optimal Labelling Designs for Sequential Decisions on Language Model Updates}
\end{center}
\medskip
\fi

\bigskip
\begin{abstract}
When a large language model (LLM) is updated, its developer must decide whether accuracy improved, worsened or stayed within a tolerance. An LLM judge can assess every prompt cheaply but imperfectly, while human labels are costly. We propose Sequential Decisions with Optimal Labelling Designs (SOLD), which labels one prompt at a time, uses the judge to correct the estimate of the change in accuracy and to choose the prompts, and stops once the evidence is conclusive. A wrong decision has probability at most a prespecified level whenever labelling stops. The commonly used Neyman allocation minimises the variance of the estimate and therefore rarely labels prompts that the judge is confident about. It can need far more draws when the outcomes of these prompts could still fall far below the judge's prediction. We instead derive the design that gathers evidence fastest, which labels each prompt with probability at least proportional to how far its outcome could fall below the prediction. As the level tends to zero, no procedure based on the judge-corrected estimates needs fewer draws than SOLD. In simulations and on the EVOUNA and MMLU-Pro datasets, SOLD needed fewer labels than the other methods.
\end{abstract}

\vfill

\newpage
\spacingset{1.8} %
\section{Introduction}\label{sec:intro}

When a large language model (LLM) is updated, its developer needs to know whether accuracy has improved, worsened or stayed unchanged up to a specified tolerance. In recursive self-improvement, LLM agents rewrite their own code or prompts and fine-tuning is scheduled continually \citep*{chen2026recursive}, so every round produces an update that has to be checked before it is kept. Human labels settle such questions, but they are expensive. An LLM used as a judge can assess the responses of both versions to every prompt cheaply, but its assessments are imperfect, so it can guide the labelling but cannot replace it. A judge that shares the biases of the signal used to develop an update can even overstate its gain \citep*{chen2026auditing}. As judges become more accurate, fewer labels are needed, and which prompts receive them matters more. The developer wants to use the judge to reach this three-way decision with as few labels as possible, while keeping the probability of a wrong decision at most a prespecified level $\alpha$.

We consider the setting in which labels are collected one at a time, so that labelling can stop as soon as the evidence is conclusive. A budget fixed in advance must be chosen before knowing how many labels are needed, which depends on how far the change in accuracy is from the tolerance, so it can waste labels or run out before the evidence is conclusive. Applying a test designed for a fixed sample size repeatedly as labels arrive, and stopping at the first rejection, inflates the type I error. A test whose type I error is at most $\alpha$ at any stopping time is called anytime valid \citep{ville1939,shafer2021testing,ramdas2023game}. Testing by betting yields such tests. A test starts with wealth one, bets against its null hypothesis at every label and rejects when its wealth reaches $1/\alpha$. If no bet is favourable under the null hypothesis and every bet keeps the wealth nonnegative, Ville's inequality bounds the probability of a false rejection by $\alpha$ however the test is stopped.

For a fixed budget of labels, prediction-powered inference \citep{angelopoulos2023ppi} estimates a mean outcome from the judge's predictions on all prompts and corrects their bias with the labels of a uniform sample. \citet{dorner2025limits} show that such a correction saves at most half of the labels when the judge is no more accurate than the model it evaluates. Active inference \citep{zrnic2024active}, the cost-optimal sampling of \citet{angelopoulos2025cost} and the method of \citet{wu2026efficient} instead label more often the prompts on which the judge is uncertain, so as to reduce the variance of the corrected estimate, and \citet{sawade2012active} choose the prompts to label so as to maximise the asymptotic power of a test that compares two models. Unlike active learning, which chooses inputs to label so as to train a predictor \citep{settles2009active}, these methods choose prompts so as to estimate a mean, and they draw them at random with known probabilities so that the estimate is unbiased. Their guarantees, however, hold for a fixed budget, and they do not cover labelling that stops at a time chosen from the data.

Anytime-valid methods have also been developed for related problems. \citet{balachandran2026pay} decide whether a model update makes the loss worse by more than a tolerance. Their estimate uses only the labels, so a judge can at most decide which inputs are labelled. \citet{park2025adaptive} test whether the risk of one model is below a threshold with the help of a judge but label prompts in random order, \citet{mondal2026old} test the absolute risk of a new model, and \citet{ao2026best} identify the best of several models with a judge. Closest to our approach, \citet{hsu2026efficient}, \citet{li2026celeus} and \citet{gao2026empirical} build confidence sequences, that is, intervals that contain the target at all times with a prespecified probability, for the accuracy or the risk of a single model, or for the mean of human labels. They weight each label by the inverse of its labelling probability and use predictions of the labels to decide which prompts to label, and \citet{hsu2026efficient} also use the predictions to correct the estimate. \citet{lin2026not} audit the labels of a fixed dataset with confidence sequences and choose which class of labels to inspect next from its remaining uncertainty and its effect on a downstream score. \citet{xu2024active} control the risk of prediction sets, which are meant to contain the correct output, with actively chosen labels, and \citet{csillag2025prediction} construct e-values, nonnegative statistics whose expectation is at most one under the null hypothesis, from predictions when the labelling probabilities are given. None of these methods makes the three-way decision about an update with anytime validity while using the judge both to correct the estimate of the change and to choose which prompts to label.

To fill this gap, we propose Sequential Decisions with Optimal Labelling Designs (SOLD). Let $\theta$ be the mean, over a pool of prompts, of an outcome that is $1$ if only the new version answers a prompt correctly, $-1$ if only the old version does and $0$ otherwise, and let $\delta$ be the tolerance. The update is improved if $\theta>\delta$, worsened if $\theta<-\delta$ and negligible if $|\theta|\le\delta$, and SOLD decides which of the three holds with four one-sided tests of $\theta$ against $\pm\delta$, as in tests of equivalence \citep{schuirmann1987comparison}. At each step SOLD draws one prompt for labelling according to the labelling probabilities, or the design, which it computes from the judge's probabilities of the three outcomes. It draws with replacement, reuses the label of a prompt that is drawn again, and lets the four tests share the draws. Each test bets on the prediction-powered estimate of $\theta$ computed from the label of the drawn prompt, so the judge enters both the estimate and the choice of the prompts. Each test rejects when its wealth reaches $2/\alpha$. For any $\theta$, a wrong decision requires the rejection of one of two null hypotheses that hold, so its probability is at most $\alpha$ however labelling is stopped (Section~\ref{sec:decisions}).

The number of draws that a test needs depends on its design. For a single test with a fixed design and a fixed bet, the logarithm of the wealth is, given the outcomes of all prompts, a sum of independent and identically distributed terms. When their mean $g$, the growth rate of the wealth, is positive, the wealth reaches its threshold after about $\log(1/\alpha)/g$ draws for small $\alpha$. We therefore choose the design whose growth rate under the best bet is largest. This is the criterion of \citet{kelly1956} and \citet{breiman1961} for repeated bets, and \citet{hsu2026efficient} used it to choose the prompts to label when estimating the accuracy of a single model. They compute this design numerically at each step and do not characterise its structure. \citet{sandoval2026multi} use the same criterion to choose which of several data sources to sample when testing a global null hypothesis, and show that the expected rejection time then matches a lower bound. \citet{spertus2024sequential} choose the stratum to sample next and the bets of a test of the mean of a stratified population so as to minimise its expected stopping time. The design with the largest growth rate can differ markedly from Neyman allocation \citep{neyman1934}, the design that minimises the variance of the estimate, which is the natural target for a fixed budget. The clearest difference comes from a constraint that the variance ignores. Every bet must keep the wealth nonnegative for every prompt that can be drawn and every possible label. The bet is chosen before the prompt is drawn, and the estimate divides the difference between the label and the judge's prediction by the labelling probability of the prompt. For example, if the judge gives probability 0.99 that only the new version answers a prompt correctly, the judge's prediction for this prompt is close to 1, but its outcome can still be $-1$, nearly 2 below the prediction. Because the judge is confident, the design that minimises the variance labels this prompt rarely. A label of $-1$ would then make the estimate far negative, so every bet must stay small, and the wealth grows slowly even if no such label ever occurs.

Our theory first characterises the optimal design. For one test, we derive the design that maximises the growth rate (Theorem~\ref{thm:char}). It labels every prompt at least in proportion to the downside of the prompt, the largest amount by which its outcome can fall below the judge's prediction. Above this floor, the design equalises across prompts the marginal gain in growth rate. A design that labels some prompt rarely relative to its downside has a small growth rate (Proposition~\ref{prop:safetybound}). Even if the outcomes follow the judge's probabilities, the growth rate of Neyman allocation can be much smaller than that of the optimal design when the judge is confident about prompts with a large downside, as in the example above (Theorem~\ref{thm:threshold}). Mixing it with uniform sampling keeps the labelling probabilities above a floor, but this floor does not grow with the downside (Section~\ref{sec:variance}).

SOLD estimates the optimal design as labels arrive. As $\alpha\to0$, a single test of SOLD attains a lower bound on the expected number of draws among tests that bet on the prediction-powered estimates and reject with probability tending to one (Theorem~\ref{thm:lb}). When the update is negligible, SOLD may choose its design at random from two designs at each step, and randomising in this way can need fewer draws than any single design (Proposition~\ref{prop:pair} and Section~\ref{sec:three}). With this, each of the three decisions of SOLD attains the corresponding lower bound (Theorem~\ref{thm:lbD}). Because a prompt drawn again keeps its label, the number of labels can be smaller than the number of draws (Section~\ref{sec:lb}).

In simulations, on updates of the question-answering systems in EVOUNA \citep{wang2023evaluating} and on updates of language models evaluated on MMLU-Pro \citep{wang2024mmlupro}, SOLD needed fewer labels in total than every other method. Its advantage grew with the accuracy of the judge in the simulations and was largest on MMLU-Pro, whereas Neyman allocation needed more labels as the judge's accuracy rose from 93.3\% to 99.9\%. In the rest of the paper, Section~\ref{sec:alg} introduces the problem and SOLD, Section~\ref{sec:char} gives its theoretical properties, Section~\ref{sec:sim} reports the simulation studies, Section~\ref{sec:realdata} applies SOLD to EVOUNA and MMLU-Pro, and Section~\ref{sec:disc} concludes with a discussion. The proofs, implementation details and additional numerical results are given in the supplement.

\section{Methodology}\label{sec:alg}

\subsection{Problem setup}\label{sec:problem}

Let $P$ be the uniform distribution on a pool of $N$ prompts. The outcome $Y(x)$ of a prompt $x$ is $1$ if only the new version of the model answers it correctly, $-1$ if only the old version does, and $0$ otherwise, and a human who labels $x$ observes $Y(x)$. Write $\E_P$ for expectation when a prompt $X$ is drawn from $P$ and $Y=Y(X)$. The target is the mean outcome $\theta=\E_P[Y]$, the change in accuracy. If the two versions give the same final answer, $Y(x)=0$ is known without a label. We write $[\ell(x),u(x)]$ for the range of $Y(x)$, so that $\ell(x)=u(x)=0$ for such a prompt and $\ell(x)=-1$, $u(x)=1$ otherwise. Because human labels need not be deterministic, we regard $Y(x)$ as a random variable with a distribution $F(\cdot\mid x)$ on $[\ell(x),u(x)]$. It is generated once and then fixed, because a prompt that is drawn again keeps its label, and $\theta$ and the results of Sections~\ref{sec:alg} and~\ref{sec:char} are for the outcomes as generated. We call a prompt with $\ell(x)<u(x)$ a candidate, and write $\pi=P\{u(X)>\ell(X)\}$ for the proportion of candidates.

Before any label is collected, the judge supplies for every prompt $x$ its probabilities $p_+(x)$, $p_-(x)$ and $p_0(x)$ of $Y(x)=1$, $-1$ and $0$, with $p_0(x)=1$ for a prompt that is not a candidate. Write $\mathbf p(x)=(p_+(x),p_-(x),p_0(x))$. The judge's predictive distribution has mean $m(x)=p_+(x)-p_-(x)$, which lies in $[\ell(x),u(x)]$, and standard deviation $\sigma(x)$, where $\sigma(x)^2=p_+(x)+p_-(x)-m(x)^2$. The mean prediction $\mu=\E_P[m(X)]$ is therefore known before labelling. We call $h(x)=m(x)-\ell(x)$, the largest amount by which $Y(x)$ can fall below $m(x)$, the downside of $x$.

For a tolerance $\delta>0$, consider the four null hypotheses
\[
H_0^{(1)}:\theta\le\delta,\qquad H_0^{(2)}:\theta\ge-\delta,\qquad H_0^{(3)}:\theta\ge\delta,\qquad H_0^{(4)}:\theta\le-\delta.
\]
The update is improved if $H_0^{(1)}$ is false, worsened if $H_0^{(2)}$ is false, and negligible otherwise. Labels are collected one at a time, and the update is declared improved when $H_0^{(1)}$ is rejected, worsened when $H_0^{(2)}$ is rejected, and negligible when both $H_0^{(3)}$ and $H_0^{(4)}$ are rejected. A decision is wrong if it declares one of the three when another holds. Since $Y(x)=0$ for a prompt that is not a candidate and $|Y(x)|\le1$ otherwise, $|\theta|\le\pi$, so the update is negligible without any label if $\delta\ge\pi$, and we assume $\delta<\pi$.

\subsection{Prediction-powered betting test}\label{sec:setting}

We first construct a test of $H_0^{(1)}$. Section~\ref{sec:decisions} combines this test with the tests of the other three null hypotheses. At step $i$ we choose a design $r_i$, a nonnegative function with $\E_P[r_i(X)]=1$, draw $X_i=x$ with probability $r_i(x)/N$ and observe its label $Y_i=Y(X_i)$. We call $r_i(x)$ the labelling probability of $x$ relative to uniform sampling. The label gives the increment
\begin{equation}\label{eq:Z}
Z_i=\mu+\frac{Y_i-m(X_i)}{r_i(X_i)},
\end{equation}
the prediction-powered estimate of $\theta$ \citep{angelopoulos2023ppi,zrnic2024active} computed from that label alone. With bets $\lambda_i\ge0$ the wealth after $n$ draws is
\begin{equation}\label{eq:W}
W_n=\prod_{i=1}^n\{1+\lambda_i(Z_i-\delta)\}.
\end{equation}
The test rejects $H_0^{(1)}$ as soon as $W_n\ge1/\alpha$. The wealth $W_n$ is a special case of the prediction-powered e-values of \citet{csillag2025prediction}. We sample with replacement and reuse the label of a prompt that is drawn again, so each label is paid for once. Write $\mathcal F_i$ for the history after $i$ draws, which consists of the prompts drawn and their labels. The design $r_i$ and the bet $\lambda_i$ should be predictable, that is, $\mathcal F_{i-1}$-measurable.

The type I error is at most $\alpha$ if, under $H_0^{(1)}$, every factor $1+\lambda_i(Z_i-\delta)$ is nonnegative and
$\E[1+\lambda_i(Z_i-\delta)\mid\mathcal F_{i-1}]\le1$. Then $W_n$ is a nonnegative supermartingale with respect to $(\mathcal F_n)_{n\ge0}$, and Ville's
inequality \citep{ville1939} gives $\Pr(W_n\ge1/\alpha\text{ for some }n)\le\alpha$ \citep{shafer2021testing,ramdas2023game,waudbysmith2024estimating}. Since $\lambda_i\ge0$ is $\mathcal F_{i-1}$-measurable and $\theta\le\delta$ under $H_0^{(1)}$, $\E[1+\lambda_i(Z_i-\delta)\mid\mathcal F_{i-1}]\le1$ if $\E[Z_i\mid\mathcal F_{i-1}]\le\theta$, which is the case when $r_i(x)>0$ for every prompt $x$ with $h(x)>0$.

To see this, fix a design $r$ and drop the index $i$. Write $\E_{Q_r}$ for expectation when $X$ is drawn from $Q_r(dx)=r(x)P(dx)$, the distribution of the drawn prompt, and $Y=Y(X)$.
Only prompts with $r(x)>0$ can be drawn, and $\mu+\E_P[Y-m(X)]=\theta$, so
\begin{equation}\label{eq:EZ}
\E_{Q_r}Z=\theta-\E_P\big[\ind\{r(X)=0\}\{Y-m(X)\}\big].
\end{equation}
If $m(x)=\ell(x)$ at every prompt with $r(x)=0$, then $Y(x)-m(x)\ge0$ at these prompts, so the expectation subtracted in~\eqref{eq:EZ} is nonnegative. We therefore call
a design eligible if $r(x)>0$ for every prompt $x$ with $h(x)>0$. Under an eligible design
$\E_{Q_r}Z\le\theta$ whatever the values of $Y(x)$, so under $H_0^{(1)}$, $\E[Z_i\mid\mathcal F_{i-1}]\le\theta\le\delta$.

The factor $1+\lambda(Z-\delta)$ is nonnegative for every possible value of $Y(x)$ at every prompt $x$ with $r(x)>0$ if and only if it
is nonnegative at $Z=\mu-H(r)$, the smallest possible value of $Z$, where $H(r)=\max_{x:\,r(x)>0}h(x)/r(x)$. We call a bet $\lambda\ge0$ safe for the design $r$ if
\begin{equation}\label{eq:safebet}
\lambda\{\delta-\mu+H(r)\}\le1 .
\end{equation}
With eligible designs and safe bets the test is anytime valid, as Proposition~\ref{prop:valid}(a) in Section~\ref{sec:decisions} states.

\subsection{Labelling design and bet}\label{sec:design}\label{sec:bets}

We measure a design $r$ by its growth rate. For distributions $\bar F(\cdot\mid x)$ on $[\ell(x),u(x)]$, let
\begin{equation}\label{eq:G}
\begin{aligned}
G_{\bar F}(r)&=\sup_{\lambda\ \mathrm{safe}}g_{\bar F}(r,\lambda),\\
g_{\bar F}(r,\lambda)&=\E_P\Big[r(X)\int\log\Big\{1+\lambda\Big(\mu-\delta+\frac{y-m(X)}{r(X)}\Big)\Big\}\,\bar F(dy\mid X)\Big],
\end{aligned}
\end{equation}
where the term of a prompt with $r(x)=0$ is taken to be zero, since such a prompt is never drawn. Then $g_{\bar F}(r,\lambda)$ is the expectation of $\log\{1+\lambda(Z-\delta)\}$ when $X$ is drawn from $Q_r$ and $Y$ from $\bar F(\cdot\mid X)$. For a design that is not eligible, let $G_{\bar F}(r)=0$. Write $G$ and $g$ for $G_{\bar F}$ and $g_{\bar F}$ with $\bar F(\cdot\mid x)$ the point mass at $Y(x)$, and call a design optimal if it attains the largest growth rate $G^\star=\sup_{r\ \mathrm{eligible}}G(r)$. Since $G$ depends on the outcomes $Y(x)$, which are unknown, at step $i$ we use $\bar F=\hat F_i$, where $\hat F_i(\cdot\mid x)$ is the point mass at $Y(x)$ if $x$ has been labelled before step $i$ and $\mathbf p(x)$ otherwise. We compute the design $r'$ that maximises $G_{\hat F_i}(r)$ and the safe bet $\lambda'$ that attains $G_{\hat F_i}(r')$.

Write the bet as $\lambda_t=t/\{1-t(\mu-\delta)\}$, where $t$ is the normalised bet. By~\eqref{eq:safebet}, $\lambda_t$ is safe for an eligible design $r$ exactly when
\begin{equation}\label{eq:safe}
r(x)\ge t\,h(x)\quad\text{for every prompt }x,
\end{equation}
and we call $t\,h(x)$ the safety floor of $x$. For a given $t$ and distributions $\bar F$, let
\begin{equation}\label{eq:design}
r_{t,\bar F}(x)=\max\Big\{\frac{t}{\Psi_{x,\bar F}^{-1}(\nu_{t,\bar F})},\ t\,h(x)\Big\},\qquad
\Psi_{x,\bar F}(s)=\E_{\bar F}\Big[\log(1+se)-\frac{se}{1+se}\,\Big|\,X=x\Big],
\end{equation}
where $e=Y-m(X)$ and $\E_{\bar F}[\,\cdot\mid X=x]$ denotes expectation with $Y$ drawn from $\bar F(\cdot\mid x)$. Here $\Psi_{x,\bar F}^{-1}$ is a generalised inverse of $\Psi_{x,\bar F}$ on $[0,1/h(x))$ and, for $0<t<1/\E_P[h(X)]$, $\nu_{t,\bar F}$ is the constant for which $\E_P[r_{t,\bar F}(X)]=1$, the Lagrange multiplier of this constraint. Supplementary~S1.1 gives both precisely. Above its safety floor, a prompt has $\Psi_{x,\bar F}\{t/r_{t,\bar F}(x)\}=\nu_{t,\bar F}$, which equalises across these prompts the marginal gain in growth rate from a larger labelling probability (Section~\ref{sec:charthm}). At $t=1/\E_P[h(X)]$ the design is $r_{t,\bar F}(x)=h(x)/\E_P[h(X)]$. Let $t^\star$ maximise $g_{\hat F_i}(r_{t,\hat F_i},\lambda_t)$ over
$0<t\le1/\E_P[h(X)]$. Then $r'=r_{t^\star,\hat F_i}$ and $\lambda'=\lambda_{t^\star}$ by Theorem~\ref{thm:char}.

We compute the design~\eqref{eq:design} by a search over $\nu_{t,\hat F_i}$ (Supplementary~S1.1). Since $\hat F_i$ changes only when a prompt is labelled for the first time, $r'$ and $\lambda'$ are recomputed only then. If the mean outcome under $\hat F_i$, $\hat\theta_i=\E_P\{\int y\,\hat F_i(dy\mid X)\}$, is at most $\delta$, then $G_{\hat F_i}(r)=0$ for every eligible design $r$, and we set $r'(x)=\ind\{u(x)>\ell(x)\}/\pi$ for every $x$ and $\lambda'=0$.

If the judge's probabilities make $r'(x)$ small for a candidate whose $Y(x)$ differs from the judge's prediction, $x$ may be labelled too rarely for this error to be corrected. We therefore mix $r'$ with uniform sampling of the candidates, as in \citet{zrnic2024active} and \citet{hsu2026efficient}, and draw $X_i$ from $r_i(x)=(1-\varepsilon_i)r'(x)+\varepsilon_i\ind\{u(x)>\ell(x)\}/\pi$ with weight $\varepsilon_i=i^{-1/2}/4$. We start from the base bet
\begin{equation}\label{eq:bet}
\hat\lambda_i=\frac{(1-\varepsilon_i)\lambda'}{1+\varepsilon_i\lambda'(\mu-\delta)},
\end{equation}
which is safe for $r_i$ by Lemma~S5. Because $\lambda'$ is the best safe bet for $G_{\hat F_i}(r')$ rather than for $G(r')$, it can be too large, and a bet larger than the best one lowers the growth rate, so a multiple $b\hat\lambda_i$ with $b<1$ may grow faster. Every multiple with $b\in[0,1]$ is safe, because the safe bets form an interval that contains 0. We therefore average the wealth over $b\in[0,1]$, as in the universal portfolio of \citet{cover1991universal},
\[
W_n=\int_0^1\prod_{i\le n}\big\{1+b\,\hat\lambda_i(Z_i-\delta)\big\}\,db .
\]
For every $b\in[0,1]$, $\log W_n\ge\log\prod_{i\le n}\{1+b\,\hat\lambda_i(Z_i-\delta)\}-\log(n+1)-1$ (Lemma~S7), so $W_n$ grows nearly as fast as the wealth of the best $b$ chosen after seeing the data.
This average equals the wealth~\eqref{eq:W} of the single predictable safe bet $\lambda_i=\bar b_i\hat\lambda_i$, where
$\bar b_i$ is the mean of $b$ under weights proportional to $\prod_{j<i}\{1+b\,\hat\lambda_j(Z_j-\delta)\}$. The test rejects at $T_\alpha=\inf\{n:W_n\ge1/\alpha\}$ (Algorithm~S1 in Supplementary~S1.1).

\begin{remark}[Refitting the judge's probabilities]\label{rem:refit}
The test needs more labels when the judge's probabilities are biased or overconfident. We can refit them to the labels collected so far, for example by the refitting model of Supplementary~S1.2, which changes the predictions $m(x)$, the mean prediction $\mu$ and the design. Each refit at step $i$ uses only $\mathcal F_{i-1}$, and the designs and bets computed from the refitted probabilities are eligible and safe, so Proposition~\ref{prop:valid}(a) applies with refitting.
\end{remark}

\subsection{Three-way decision}\label{sec:decisions}\label{sec:errorcontrol}\label{sec:together}

With $-\delta$ in place of $\delta$, the test of Sections~\ref{sec:setting} and~\ref{sec:design} tests $H_0^{(4)}:\theta\le-\delta$. With $-Y$, $-u(x)$, $-\ell(x)$ and $-m(x)$ in place of $Y$, $\ell(x)$, $u(x)$ and $m(x)$, it tests $H_0^{(2)}:-\theta\le\delta$, and $H_0^{(3)}:-\theta\le-\delta$ if $\delta$ is also replaced by $-\delta$.

We let the four tests share the draws. Write $W_n^{(k)}$ for the wealth of the test of $H_0^{(k)}$ after $n$ draws, $k=1,\dots,4$, and add to $\mathcal F_n$ the randomisation used to choose the designs and the bets of the first $n$ steps. We use one design $r'$ at each step, chosen for the decision that $\hat\theta_i$ indicates. If $\hat\theta_i>\delta$, $r'$ is the design of the test of $H_0^{(1)}$. If $\hat\theta_i<-\delta$, it is that of the test of $H_0^{(2)}$. While $|\hat\theta_i|\le\delta$, both $H_0^{(3)}$ and $H_0^{(4)}$ must be rejected, so $r'$ must serve both tests, although they bet in opposite directions. Write $G_{k,\hat F_i}(r)$ for the growth rate~\eqref{eq:G} of the test of $H_0^{(k)}$, computed under $\hat F_i$. For $w\in[0,1]$, Lemma~S14 in Supplementary~S3.5 gives the form of the designs that maximise $wG_{3,\hat F_i}(r)+(1-w)G_{4,\hat F_i}(r)$. Let $w^\star$ minimise $\max_r\{wG_{3,\hat F_i}(r)+(1-w)G_{4,\hat F_i}(r)\}$ over $w\in[0,1]$, and let $\hat\Gamma$ be this minimum. If some maximiser $r$ at $w^\star$ has $\min\{G_{3,\hat F_i}(r),G_{4,\hat F_i}(r)\}=\hat\Gamma$, as when $G_{3,\hat F_i}(r)=G_{4,\hat F_i}(r)$, we take $r'=r$. Otherwise we take two maximisers $r_a$ and $r_b$ at $w^\star$ and set $r'=r_a$ with probability $\varrho$ and $r'=r_b$ otherwise, where $\varrho G_{3,\hat F_i}(r_a)+(1-\varrho)G_{3,\hat F_i}(r_b)=\varrho G_{4,\hat F_i}(r_a)+(1-\varrho)G_{4,\hat F_i}(r_b)$ (Proposition~\ref{prop:pair}). Once exactly one of the tests of $H_0^{(3)}$ and $H_0^{(4)}$ has rejected, and while $|\hat\theta_i|\le\delta$, $r'$ is the design of the other.

We draw prompts from $(1-\varepsilon_i)r'(x)+\varepsilon_i\ind\{u(x)>\ell(x)\}/\pi$ as in Section~\ref{sec:design}. Every test that has not rejected bets on $Z_i$, or on $-Z_i$ if it tests $-Y$, with the base bet~\eqref{eq:bet}. For the test of $H_0^{(k)}$, $\lambda'$ in~\eqref{eq:bet} attains $G_{k,\hat F_i}(r')$, or is $0$ if $r'$ is not eligible for that test. The design in use is eligible for every test because it is mixed with uniform sampling of the candidates, and every base bet is safe for its test. Hence, for every $k$ for which $H_0^{(k)}$ holds, $W^{(k)}$ is a nonnegative supermartingale with respect to $(\mathcal F_n)$, by the argument of Section~\ref{sec:setting} applied given $\mathcal F_{i-1}$ and the randomisation of step $i$, also with refitted probabilities.

The test of $H_0^{(k)}$ rejects when $W_n^{(k)}\ge2/\alpha$. For any $\theta$, a wrong decision requires the rejection of one of two null hypotheses that hold, so this threshold keeps the probability of a wrong decision at most $\alpha$. Labelling stops at the first decision, or earlier with no decision. If every candidate has been labelled before a decision, $\theta$ is known and the decision is read off. Algorithm~S2 in Supplementary~S1.1 summarises the procedure. Supplementary~S2 proves all results in the order in which they appear.

\begin{proposition}\label{prop:valid}
Fix the outcomes $Y(x)$ of all prompts, so that the only randomness is in the draws $X_i$ and in the choice of designs.
\begin{enumerate}
\item[(a)] Suppose that the predictions $m(x)$, the design $r_i$ and the bet $\lambda_i$ used at step $i$ are
$\mathcal F_{i-1}$-measurable, that each design is eligible and that each bet is
safe for its design. If $H_0^{(1)}$ holds, $\Pr(W_n\ge1/\alpha\text{ for some }n)\le\alpha$.
\item[(b)] Suppose that, for every $k$ for which $H_0^{(k)}$ holds, $(W_n^{(k)})_{n\ge0}$ is a nonnegative supermartingale with respect to $(\mathcal F_n)_{n\ge0}$ with $W_0^{(k)}=1$. Then, with probability at least $1-\alpha$, no wrong decision is made, however labelling is stopped.
\end{enumerate}
\end{proposition}

\section{Theoretical analysis}\label{sec:char}

Section~\ref{sec:charthm} derives the design that maximises the growth rate. Section~\ref{sec:opt} shows that the largest growth rate determines how many draws a test that bets on $Z_i$ needs, and that our test needs, to first order as $\alpha\to0$, the fewest. Section~\ref{sec:three} extends these results to the three decisions, and Section~\ref{sec:variance} compares the optimal design with Neyman allocation. 

\subsection{Optimal design}\label{sec:charthm}

Write $e=Y-m(X)$ for the residual of a drawn prompt, $\Delta=\mu-\delta$, which we call the gap, and $t=\lambda/(1+\lambda\Delta)$ for the normalised bet of Section~\ref{sec:design}. Since $Z-\delta=\Delta+e/r(X)$, we have $1+\lambda(Z-\delta)=(1+\lambda\Delta)\{1+t\,e/r(X)\}$ and $1+\lambda\Delta=(1-t\Delta)^{-1}$, so, for distributions $\bar F$ as in Section~\ref{sec:design},
\begin{equation}\label{eq:growth}
g_{\bar F}(r,\lambda)=-\log(1-t\Delta)-C_{\bar F}(r,t),\qquad C_{\bar F}(r,t)=\E_P\big[r(X)\,c_{X,\bar F}\big(t/r(X)\big)\big],
\end{equation}
where $c_{x,\bar F}(s)=-\E_{\bar F}[\log(1+se)\mid X=x]$, and the term of a prompt with $r(x)=0$ is again taken to be zero. The design enters only through $C_{\bar F}(r,t)$, which we call the cost of the design $r$ at the bet $t$. The cost is the expected loss in the logarithm of the wealth due to the residuals. Write $C_{\bar F}(t)=C_{\bar F}(r_{t,\bar F},t)$ for $0<t\le1/\E_P[h(X)]$, $\sigma_{\bar F}(x)^2=\E_{\bar F}[e^2\mid X=x]$ and $\theta_{\bar F}=\E_P\{\E_{\bar F}[Y\mid X]\}$.

\begin{theorem}[Optimal design]\label{thm:char}
Suppose that $\E_P[h(X)]>0$ and $P\{\sigma_{\bar F}(X)>0\}>0$.
\begin{enumerate}
\item[(a)] For $0<t<1/\E_P[h(X)]$, the design $r_{t,\bar F}$ is the unique minimiser of the cost $C_{\bar F}(r,t)$ over eligible designs satisfying~\eqref{eq:safe}.
\item[(b)] If $\theta_{\bar F}>\delta$, then $r_{t^\star,\bar F}$ maximises $G_{\bar F}(r)$ over eligible designs, where $t^\star$ maximises $-\log(1-t\Delta)-C_{\bar F}(t)$ over $0<t\le1/\E_P[h(X)]$.
\end{enumerate}
\end{theorem}

In Theorem~\ref{thm:char}(a), the design minimises the cost $C_{\bar F}(r,t)$ subject to $\E_P[r(X)]=1$ and the floors $r(x)\ge t\,h(x)$. Prompt $x$ contributes $r(x)\,c_{x,\bar F}\{t/r(x)\}$ to the cost, through the effective bet $s=t/r(x)$ on its residual. Since $\Psi_{x,\bar F}(s)=s\,c_{x,\bar F}'(s)-c_{x,\bar F}(s)$, this contribution has derivative $-\Psi_{x,\bar F}\{t/r(x)\}$ with respect to $r(x)$, and since $\Psi_{x,\bar F}$ increases from $\Psi_{x,\bar F}(0)=0$ (Lemma~S1), it decreases in $r(x)$ and is convex, that is, labelling a prompt more often lowers its contribution, by less and less. The minimiser therefore has $\Psi_{x,\bar F}\{t/r(x)\}=\nu_{t,\bar F}$, the same value, for every prompt above its floor, which gives $r(x)=t/\Psi_{x,\bar F}^{-1}(\nu_{t,\bar F})$ in~\eqref{eq:design}. With $\bar F(\cdot\mid x)$ the point mass at $Y(x)$, (b) gives an optimal design.

\subsection{Lower bound and achievability}\label{sec:opt}\label{sec:lb}\label{sec:ach}\label{sec:relation}

We now give a lower bound on the expected number of draws of any test that bets on $Z_i$ in~\eqref{eq:Z}, and show that our test attains it. A test chooses the design $r_i$ from $\mathcal F_{i-1}$ and draws $X_i$ from it, and the labels it sees form a process $(X_i,Y_i)_{i\ge1}$. For the pool of Section~\ref{sec:problem}, $Y_i=Y(X_i)$, and we write $Q$ for the distribution of this process, under which, given $\mathcal F_{i-1}$ and $r_i$, $X_i$ has the distribution $Q_{r_i}$ of Section~\ref{sec:setting}. Let $\cN$ be the set of processes under which, for every $i$,
$\E[Z_i\mid\mathcal F_{i-1},r_i]\le\delta$ and $Z_i\ge\mu-H(r_i)$, the smallest possible value of $Z_i$ (Section~\ref{sec:setting}). It contains the processes with $Y_i=Y(X_i)$ in which the designs are eligible and $H_0^{(1)}$ holds. When the labels come from a process in $\cN$, the wealth of a test that bets on $Z_i$ with predictable designs and safe bets is a nonnegative
supermartingale, so the test rejects with probability at most $\alpha$. Write $T_\alpha$ for the number of draws after which a test at level $\alpha$ stops.

\begin{theorem}[Lower bound and achievability]\label{thm:lb}
Suppose that $\theta>\delta$. In (b) and (c), suppose in addition that $\E_P[h(X)]>0$ and that the judge's probabilities are not refitted (Remark~\ref{rem:refit}).
\begin{enumerate}
\item[(a)] Consider a sequential test that draws prompts with replacement, chooses eligible designs predictably and has type I error
at most $\alpha$ under every process in $\cN$. If it rejects $H_0^{(1)}$ with probability at least $1-\varpi$ under $Q$, for some $\varpi\in[0,1)$, then
\[
\E_Q[T_\alpha]\ge\frac{(1-\varpi)\log(1/\alpha)-\log2}{G^\star}.
\]
If the test draws from the same design $r$ at every step, the bound holds with $G(r)$ in place of $G^\star$.
\item[(b)] For a fixed eligible design $r$ with $G(r)>0$ and the largest safe bet as the base bet, $\E_Q[T_\alpha]\le\log(1/\alpha)/G(r)+O\big(\{\log(1/\alpha)\log\log(1/\alpha)\}^{1/2}\big)$ as $\alpha\to0$.
\item[(c)] For our test of Section~\ref{sec:design}, $\E_Q[T_\alpha]<\infty$ for every $\alpha$ and
\[
\lim_{\alpha\to0}\frac{\E_Q[T_\alpha]}{\log(1/\alpha)}=\frac1{G^\star},\qquad\limsup_{\alpha\to0}\frac{T_\alpha}{\log(1/\alpha)}\le\frac1{G^\star}\ \text{almost surely}.
\]
\end{enumerate}
\end{theorem}

By Theorem~\ref{thm:lb}(a), a test that is valid under every process in $\cN$ and has power close to one needs at least about $\log(1/\alpha)/G^\star$ draws. The bound comes from the least favourable process in $\cN$, which need not be a pool with fixed outcomes. For each design, $G(r)$ is the information per draw against this process \citep{honda2010asymptotically,agrawal2021optimal}. The proof follows the change-of-measure arguments of \citet{chernoff1959sequential},
\citet{kaufmann2016complexity} and \citet{garivier2016optimal}. \citet{shekhar2025optimal} proves a bound of this form for tests of a composite null hypothesis from independent observations whose distribution does not depend on a design, \citet{naghshvar2013active} give such bounds for sequential tests that choose which experiment to perform, and \citet{ham2026human} for a test of a proportion with scores from an AI system and human labels.

By part (a), the limit $1/G^\star$ in (c) is the smallest that a test whose power tends to one can attain, so our test is asymptotically optimal among tests that bet on $Z_i$. Part (c) also holds when the design and the bet are computed only approximately and for other weights $\varepsilon_i$ (Theorem~S2 in Supplementary~S3.1). The test in (b) has $\E_Q[T_\alpha]<\infty$, so it rejects with probability one, and part (a) with $G(r)$ in place of $G^\star$ and part (b) give $\E_Q[T_\alpha]/\log(1/\alpha)\to1/G(r)$. A design $r$ therefore needs about $G^\star/G(r)$ times as many draws as an optimal design.

Theorem~\ref{thm:lb} counts draws, but only labels are paid for. In SOLD a prompt that is drawn again needs no new label, so the number of labels is at most the number of draws, and the upper bounds of Theorem~\ref{thm:lb}(b) and (c) also bound the expected number of labels. Proposition~S2 in Supplementary~S3.4 shows that, if every prompt is a candidate and $r_i(x)\le r_{\max}$ for a constant $r_{\max}$, the expected number of labels is at least the lower bound of Theorem~\ref{thm:lb}(a) minus a term that is small when the pool is large compared with the number of draws needed.

\subsection{Optimality for the three decisions}\label{sec:three}

We now extend the results of Section~\ref{sec:opt} to the three decisions of Section~\ref{sec:decisions}. For $k=1,\dots,4$, let $h_k$ be the downside of the test of $H_0^{(k)}$ and $G_k(r)$ its growth rate $G(r)$, which uses $Y$ or $-Y$ and $\delta$ or $-\delta$ as in Section~\ref{sec:decisions}, so that $G_1=G$ and $G_k(r)=0$ if $r(x)=0$ for some prompt $x$ with $h_k(x)>0$. When $\theta\ne\pm\delta$, let $D$ be the correct decision, and let $K(D)$ be the set of tests that must all reject for $D$ to be declared, which is $\{1\}$ if $D$ is improved, $\{2\}$ if $D$ is worsened and $\{3,4\}$ if $D$ is negligible. The analogue of $G^\star$ for the decision $D$ is $\Gamma_D$, defined, for weights $w=(w_k)_{k\in K(D)}$ that are nonnegative and sum to one, by
\[
V_D(w)=\sup_r\sum_{k\in K(D)}w_kG_k(r),\qquad \Gamma_D=\min_wV_D(w),
\]
the supremum being over designs. If $D$ is improved, $\Gamma_D=G^\star$, and if $D$ is worsened, $\Gamma_D=\sup_rG_2(r)$. If $D$ is negligible, $\Gamma_D\ge\sup_r\min\{G_3(r),G_4(r)\}$, and $\Gamma_D$ can be larger because $G_k$ need not be concave in the design. A procedure can then do better by choosing its design at random. \citet{hsu2026efficient} face the same choice for the two tests at the end points of a confidence sequence and maximise the smaller of the two growth rates over single designs. In the two-prompt example of Supplementary~S3.6, a random choice of design needs, to first order as $\alpha\to0$, about 0.83 times as many draws as the best single design. Write $\cN_k$ for the set $\cN$ of Section~\ref{sec:opt} for the test of $H_0^{(k)}$.

\begin{proposition}[Randomised designs]\label{prop:pair}
Suppose that $D$ is negligible and that $\E_P[h_k(X)]>0$ for $k=3,4$. If $w^\star$ attains $\Gamma_D$, there are designs $r_a$ and $r_b$, possibly equal, that maximise $w_3^\star G_3(r)+w_4^\star G_4(r)$, and a probability $\varrho$, such that choosing the design $r_a$ with probability $\varrho$ and $r_b$ otherwise gives each of the two tests the expected growth rate $\varrho G_k(r_a)+(1-\varrho)G_k(r_b)\ge\Gamma_D$. No random choice of designs gives both tests an expected growth rate larger than $\Gamma_D$.
\end{proposition}

\begin{theorem}[Lower bound and achievability for the decisions]\label{thm:lbD}
Suppose that $\theta\ne\pm\delta$. In (b), suppose in addition that $\E_P[h_k(X)]>0$ for $k=1,\dots,4$ and that the judge's probabilities are not refitted.
\begin{enumerate}
\item[(a)] Consider a sequential procedure that draws prompts with replacement and chooses predictably, possibly with independent randomisation, designs that are eligible for every test in $K(D)$. Suppose that, for every $k\in K(D)$, it declares $D$ with probability at most $\alpha$ under every process in $\cN_k$, and that it declares $D$ with probability at least $1-\varpi$ under $Q$. Then
\[
\E_Q[T_\alpha]\ge\frac{(1-\varpi)\log(1/\alpha)-\log2}{\Gamma_D}.
\]
\item[(b)] For SOLD without the read-off of Section~\ref{sec:decisions} and with $r'$ and $\lambda'$ computed exactly under $\hat F_i$, $\E_Q[T_\alpha]<\infty$ for every $\alpha$ and
\[
\lim_{\alpha\to0}\frac{\E_Q[T_\alpha]}{\log(1/\alpha)}=\frac1{\Gamma_D}.
\]
\end{enumerate}
\end{theorem}

Part (a) extends Theorem~\ref{thm:lb}(a) to the three decisions, with $\Gamma_D$ in place of $G^\star$. SOLD in (b) declares $D$ only when every test in $K(D)$ has crossed its threshold $2/\alpha$. Under every process in $\cN_k$ the wealth of the test of $H_0^{(k)}$ is a nonnegative supermartingale, so by Ville's inequality it crosses $2/\alpha$ with probability at most $\alpha/2$, and SOLD satisfies the conditions of part (a). The two parts show that, for each of the three decisions, SOLD needs to first order the smallest expected number of draws among procedures that bet on $Z_i$ and $-Z_i$ and declare the correct decision with probability tending to one. If no single design attains $\Gamma_D$, which can happen only when $D$ is negligible, SOLD attains it through the random choice between two designs of Proposition~\ref{prop:pair}. Part (b) also holds when $r'$ and $\lambda'$ are computed only approximately and for other weights $\varepsilon_i$ (Theorem~S4 in Supplementary~S3.2).

\subsection{Comparison with Neyman allocation}\label{sec:variance}\label{sec:threshold}

We now compare the optimal design with Neyman allocation, the design that minimises the variance of the estimate of $\theta$. Several methods for a fixed budget in Section~\ref{sec:intro} choose their designs to make this variance small. With $n$ draws from a fixed design $r$, the estimate $n^{-1}\sum_{i=1}^nZ_i$ has variance $\Var_{Q_r}(Z)/n$. Both designs are computed from the judge's probabilities, so we compare them when these probabilities are correct, that is, when among the prompts with the same $\mathbf p(x)$ the proportions of $Y=1$, $-1$ and $0$ are $\mathbf p(x)$. We compute the growth rate and $\Var_{Q_r}(Z)$ under the judge's distribution $\bar F(\cdot\mid x)$, which puts probabilities $p_+(x)$, $p_-(x)$ and $p_0(x)$ on $1$, $-1$ and $0$, and write a subscript $\mathbf p$ in place of $\bar F$, as in $G_{\mathbf p}$ and $G^\star_{\mathbf p}=\sup_{r\ \mathrm{eligible}}G_{\mathbf p}(r)$. The design of Theorem~\ref{thm:char} attains $G^\star_{\mathbf p}$ and, like Neyman allocation and the other designs in this subsection, depends on $x$ only through $\mathbf p(x)$ and $[\ell(x),u(x)]$. For such a design $r$, $G(r)=G_{\mathbf p}(r)$ (Lemma~S3 in Supplementary~S2.8.3). Under $\mathbf p$, $\mu=\theta$, $\Delta=\theta-\delta$ and $\Var_{Q_r}(Z)=\E_P[\sigma(X)^2/r(X)]$. If $\E_P[\sigma(X)]>0$, then, by the Cauchy--Schwarz inequality and $\E_P[r(X)]=1$, $\Var_{Q_r}(Z)$ is minimised by Neyman allocation $\rN(x)=\sigma(x)/\E_P[\sigma(X)]$ \citep{neyman1934}, with minimum $\{\E_P[\sigma(X)]\}^2$.

The bet $t$ is limited in two ways, by the cost in the growth rate~\eqref{eq:growth} and by the safety condition~\eqref{eq:safe}, and the designs that maximise the growth rate and minimise $\Var_{Q_r}(Z)$ differ most when the safety condition determines the bet. The cost $C_{\mathbf p}(r,t)$ is nonnegative by Jensen's inequality, because $\E_{\mathbf p}[e\mid X=x]=0$, and increases with
$t$, and for small $t$ it is approximately $\tfrac12t^2\E_P[\sigma(X)^2/r(X)]$, which Neyman allocation minimises. If $\sigma(x)$ is bounded away from zero on the candidates, the design of Theorem~\ref{thm:char}(a), which minimises this cost subject to~\eqref{eq:safe}, tends to Neyman allocation uniformly over the prompts as $t\to0$ (Corollary~S2 in Supplementary~S3.8). The safety condition~\eqref{eq:safe} bounds the bet by $\min_{x:\,h(x)>0}r(x)/h(x)$. For $\rN$,~\eqref{eq:safe} reads $\sigma(x)/\E_P[\sigma(X)]\ge t\,h(x)$ for every $x$, that is, $t\le1/\{\kapN\E_P[\sigma(X)]\}$, where $\kapN:=\max_xh(x)/\sigma(x)$, with $h(x)/\sigma(x)=\infty$ if $\sigma(x)=0<h(x)$ and $0$ if $h(x)=\sigma(x)=0$.

Neyman allocation can give low labelling probabilities to the prompts that limit the bet most. Consider two candidates, with $\ell(x)=-1$ and $h(x)=m(x)+1$. For the first the judge gives $p_+(x)=0.99$ and $p_-(x)=p_0(x)=0.005$, so that $\sigma(x)=0.157$ and $h(x)=1.985$, and for the second $p_+(x)=p_-(x)=0.4$, so that $\sigma(x)=0.894$ and $h(x)=1$. Neyman allocation labels the first prompt about a sixth as often as the second. A prompt enters the condition~\eqref{eq:safebet} for a safe bet through $h(x)/\rN(x)=\{h(x)/\sigma(x)\}\E_P[\sigma(X)]$, and $h(x)/\sigma(x)$ is 12.6 for the first prompt and 1.1 for the second. The first prompt thus restricts the bet about eleven times as much as the second, although $Y(x)=-1$ only with probability 0.005. As the judge becomes certain of an outcome other than $-1$, $h(x)/\sigma(x)$ grows without bound.

\begin{proposition}[Safety bound]\label{prop:safetybound}
Let $\bar F$ be distributions with $\theta_{\bar F}>\delta$. For every eligible design $r$, $H(r)-\Delta\ge\delta-\E_P[\ell(X)]>0$ and
\[
G_{\bar F}(r)\le\frac{\theta_{\bar F}-\delta}{H(r)-\Delta}.
\]
\end{proposition}

Proposition~\ref{prop:safetybound} shows that a design that labels some prompt rarely relative to its downside has a small growth rate, whatever $\bar F$. The bound depends on $\bar F$ only through $\theta_{\bar F}$, and on the design only through $H(r)$. If Neyman allocation $\rN$, computed from the judge's $\sigma(x)$, is eligible, then $H(\rN)=\kapN\E_P[\sigma(X)]$, so $G_{\bar F}(\rN)\le(\theta_{\bar F}-\delta)/\{\kapN\E_P[\sigma(X)]-\Delta\}$. If $\sigma(x)/\E_P[\sigma(X)]$ tends to zero on prompts whose downside $h(x)$ stays bounded away from zero, as when the judge becomes confident of an outcome other than $-1$ on some prompts but not on all, then $\kapN\E_P[\sigma(X)]\to\infty$ and $G_{\bar F}(\rN)\to0$ for every $\bar F$, whether or not the judge's probabilities are correct. A design with $r(x)\ge t\,h(x)$ for every prompt, such as that of Theorem~\ref{thm:char}, has $H(r)\le1/t$, so Proposition~\ref{prop:safetybound} does not force its growth rate to zero however confident the judge is.

To measure how much the safety condition reduces the growth rate, we compare with the growth rate that a variance criterion predicts. When $t\Delta$ and the effective bets $t/r(x)$ are small, $g_{\mathbf p}$ in~\eqref{eq:growth} is to leading order $t\Delta-\tfrac12t^2\Var_{Q_r}(Z)$, which is maximised at $t=\Delta/\Var_{Q_r}(Z)$ with value $\Delta^2/\{2\Var_{Q_r}(Z)\}$. To this order the best design is Neyman allocation,
with bet and growth rate
\[
t_{\rm id}=\frac{\Delta}{\{\E_P[\sigma(X)]\}^2},\qquad \Gid=\frac{\Delta^2}{2\{\E_P[\sigma(X)]\}^2},
\]
which we call ideal because they ignore the safety condition, and we express growth rates as fractions of $\Gid$. Because $\Gid$ comes from a second-order expansion, $G^\star_{\mathbf p}$ can exceed it.

Two ratios measure how much safety restricts the bet. The ratio of $t_{\rm id}$ to the largest safe bet for $\rN$, $\rhoN:=t_{\rm id}\,\kapN\,\E_P[\sigma(X)]=\Delta\,\kapN/\E_P[\sigma(X)]$, measures how much safety restricts Neyman allocation, since $t_{\rm id}$ is safe for $\rN$ if and only if $\rhoN\le1$. Every design requires $t\le1/\E_P[h(X)]$, as taking expectations in~\eqref{eq:safe} with $\E_P[r(X)]=1$ shows, and the corresponding ratio $\rho^\star:=t_{\rm id}\,\E_P[h(X)]=\Delta\,\E_P[h(X)]/\{\E_P[\sigma(X)]\}^2$ measures how much safety restricts any design. Since $\E_P[h(X)]=\E_P[\{h(X)/\sigma(X)\}\sigma(X)]\le\kapN\E_P[\sigma(X)]$,
$\rho^\star\le\rhoN$, and since $h(x)\le\bar R:=\sup_x\{u(x)-\ell(x)\}$, $\rho^\star\le\bar\rho:=\Delta\bar R/\{\E_P[\sigma(X)]\}^2$.

Theorem~\ref{thm:threshold} bounds the growth rates of Neyman allocation and of the optimal design through these ratios and the functions
\begin{gather*}
\varphi(z):=\sup_{0\le k<1/z}\Big\{2k-\frac{k^2}{1-kz}\Big\}=\frac{2}{z}\cdot\frac{1-v}{1+v},\qquad
v=(1+2z)^{-1/2},\\
\varphi^\star(z):=\sup_{0<q<1}(1-q)\,\varphi\Big(\frac{(1-q)z}{q}\Big),
\end{gather*}
with $\varphi(\infty)=0$. Both equal 1 at $z=0$, decrease, and behave like $2/z$ as $z\to\infty$.

\begin{theorem}[Growth rate of Neyman allocation]\label{thm:threshold}
Suppose that the judge's probabilities are correct, that $\E_P[\sigma(X)]>0$ and that $\theta>\delta$.
\begin{enumerate}
\item[(a)] For Neyman allocation,
\[
\varphi(\rhoN)\le\frac{G_{\mathbf p}(\rN)}{\Gid}\le\frac{-\log(1-2\Gid/\rhoN)}{\Gid}=\frac{2}{\rhoN}\{1+O(\Gid/\rhoN)\}.
\]
\item[(b)] For the optimal design,
\[
\varphi^\star(\rho^\star)\le\frac{G^\star_{\mathbf p}}{\Gid}\le\frac{-\log(1-2\Gid/\rho^\star)}{\Gid}=\frac{2}{\rho^\star}\{1+O(\Gid/\rho^\star)\},
\]
and $G^\star_{\mathbf p}/\Gid\to1$ when $\bar\rho\to0$.
\end{enumerate}
\end{theorem}

A design $r$ needs about $G^\star/G(r)$ times as many draws as the optimal design (Section~\ref{sec:opt}), where $G^\star$ is the largest growth rate over all designs, including those that use information beyond $\mathbf p(x)$. Since $G(\rN)=G_{\mathbf p}(\rN)$ and $G^\star\ge G^\star_{\mathbf p}$, dividing the lower bound in (b) by the upper bound in (a) shows that Neyman allocation needs at least
\[
\frac{G^\star}{G(\rN)}\ge\frac{G^\star_{\mathbf p}}{G_{\mathbf p}(\rN)}\ge\frac{\varphi^\star(\rho^\star)\,\Gid}{-\log(1-2\Gid/\rhoN)}=\frac{\varphi^\star(\rho^\star)\,\rhoN}{2}\{1+O(\Gid/\rhoN)\}
\]
times as many draws as the optimal design, to first order as $\alpha\to0$. If $\Delta$ stays fixed and $\sigma(x)/\E_P[\sigma(X)]$ tends to zero on prompts whose downside $h(x)$ stays bounded away from zero, then $\rhoN\to\infty$ while $\rho^\star$ can stay bounded, and then this lower bound grows linearly in $\rhoN$ and $G_{\mathbf p}(\rN)/G^\star_{\mathbf p}\to0$. If $\rhoN\to0$ and $\bar\rho\to0$, then $G_{\mathbf p}(\rN)/G^\star_{\mathbf p}\to1$. For one pool of Supplementary~S3.9, as the accuracy of the judge rises from 59.9\% to 99.9\%, $\rhoN$ rises from 0.6 to about 1{,}300 while $\rho^\star$ stays below 5, and $G^\star_{\mathbf p}/G_{\mathbf p}(\rN)$ rises from 1 to about 220 (Table~S1).

Neyman allocation fails here because it labels some prompts too rarely. \citet{zrnic2024active} and \citet{hsu2026efficient} mix the design with uniform sampling, which keeps the labelling probability of every candidate away from zero. Mixing Neyman allocation with uniform sampling of the candidates with weight $\tau\in(0,1)$ gives
\begin{equation}\label{eq:rf}
r_\tau(x)=(1-\tau)\,\rN(x)+\tau\,\frac{\ind\{u(x)>\ell(x)\}}{\pi},
\end{equation}
and $\tau=1/2$ is the default of \citet{zrnic2024active} for sequential sampling. The mixture labels every candidate with probability at least $\tau/\pi$, but this floor does not grow with the downside $h(x)$, as the safety floor $t\,h(x)$ of the optimal design does. Proposition~S3 in Supplementary~S3.7 studies pools in which the judge is certain of the outcomes of some prompts, so that $\sigma(x)=0$ for them. As the gap and the proportion of the other prompts tend to zero, an asymptotically optimal design has a closed form and labels the prompts of which the judge is certain at their safety floors $t\,h(x)$, whereas Neyman allocation never labels them and has $G(\rN)=0$ when $h(x)>0$ for some of them, and the mixture with $\tau=1/2$ can need nearly twice as many draws as the optimal design.

\section{Simulation studies}\label{sec:sim}\label{sec:num}

\subsection{Data generation and comparison methods}\label{sec:simdesign}

Each simulated pool has $N=3{,}000$ prompts in two groups. The first group consists of 85\% of the prompts, all with $Y(x)=0$. In the second group, $Y(x)=0$ on about 10\% of its prompts, and the numbers with $Y(x)=1$ and $Y(x)=-1$ among its other prompts are chosen so that $\theta$ is exactly 0, 0.07, 0.09 or 0.11. We use $\delta=0.05$ and $\alpha=0.05$, so the update is negligible at $\theta=0$ and improved at the other three values, which exceed $\delta$ by 0.02, 0.04 and 0.06. Replacing $Y(x)$ by $-Y(x)$ and exchanging $p_+(x)$ and $p_-(x)$ turns an improved update into a worsened one, and every method treats the two in the same way, so we do not simulate worsened updates. In setting A the two versions give the same final answer on the first group and different final answers on the second, so the candidates are the prompts of the second group. In setting B the final answers of the two versions differ on every prompt, as when they are free text, so every prompt is a candidate.

For each candidate $x$, the judge sees a score drawn from the normal distribution with mean $\gamma Y(x)$ and variance 1, and turns it into $p_+(x)$, $p_-(x)$ and $p_0(x)$, the probabilities that $Y(x)=1$, $-1$ and $0$ given the score, by Bayes' rule with the proportions of these outcomes among the candidates as the prior. The judge's probabilities are therefore correct on average over the generation of the pool. A larger $\gamma$ separates the scores of the three outcomes more. For a candidate with $Y(x)\ne0$, the score has the sign of $Y(x)$ with probability $\Phi(\gamma)$, where $\Phi$ is the standard normal distribution function, and we call $\Phi(\gamma)$ the accuracy of the judge. We use $\gamma=1,1.5,2,2.5,3$, that is, accuracies of 84.1\%, 93.3\%, 97.7\%, 99.4\% and 99.9\%.

We compare SOLD with five other designs and one other method. We use the decision rule and the bets of Section~\ref{sec:together} for all five designs, so the four that use the judge differ from SOLD only in the design. They are PPI, uniform sampling with the judge, with $Z_i$ as in~\eqref{eq:Z} \citep{angelopoulos2023ppi}; Neyman, Neyman allocation, which is the cost-optimal rule of \citet{angelopoulos2025cost} when only labels have a cost, as in our setting; Mixture, the mixture~\eqref{eq:rf} with $\tau=1/2$ \citep{zrnic2024active}; SLS, the design of Sawade, Landwehr and Scheffer~\citeyearpar{sawade2012active}; and Uniform, uniform sampling without the judge, with $0$ in place of $m(x)$ and $\mu$ in~\eqref{eq:Z}. The other method, DISCERN \citep{balachandran2026pay}, reaches the decision in its own way and labels every candidate that it draws. We apply the refitting of Remark~\ref{rem:refit} also to PPI, Neyman, Mixture and SLS, and run SOLD and these four designs both with and without it. In every method, as in SOLD, we reuse the label of a prompt that is drawn again, so we compare methods by the number of prompts they label.

The two settings, five accuracies and four values of $\theta$ give 40 combinations. For each, we generate 100 pools and run every method once on each pool. Because wrong decisions are rare, we also set $\theta=\delta$, the largest $\theta$ at which the update is negligible and where a wrong decision is most likely, and generate 500 pools for each setting and accuracy. Section~\ref{sec:simwrongdec} reports the wrong decisions in these runs, and Section~\ref{sec:simresults} the numbers of labels in the 40 combinations. Section~\ref{sec:simrobust} repeats the 40 combinations with the judge's probabilities wrong in eight ways, with $N=1{,}000$ or $9{,}000$, and with $\delta=0.03$ or $0.07$ and $\theta$ equal to 0, $\delta+0.02$, $\delta+0.04$ or $\delta+0.06$. Supplementary~S5.1 gives the details.

\subsection{Wrong decisions}\label{sec:simwrongdec}

Every method kept the proportion of wrong decisions below $\alpha$. With $\theta=0$, 0.07, 0.09 and 0.11, no method made a wrong decision in any run. With $\theta=\delta$, every method made a wrong decision in at most 4 of the 500 runs for each setting and accuracy (Table~S6). Over the 5{,}000 runs of each method, SOLD made 1 wrong decision, SOLD with refitting 8, and the other methods at most 13. Wrong decisions were this rare because, with $\theta=\delta$, most runs labelled every candidate before reaching a decision, and the decision was then read off from $\theta$.

\subsection{Number of labels}\label{sec:simresults}

\begin{figure}[t]
\centering
\includegraphics[width=\linewidth]{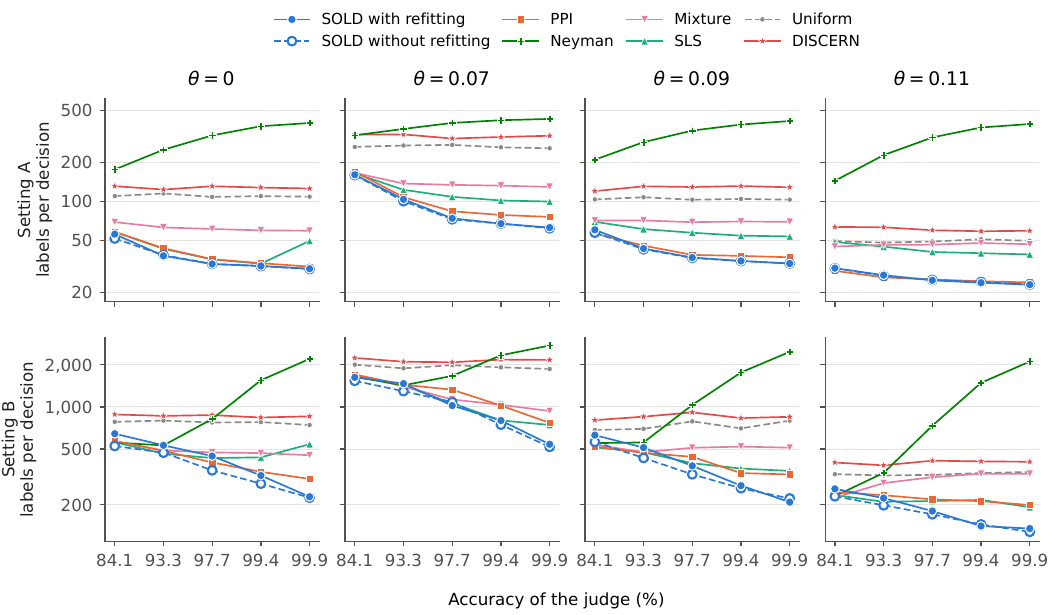}
\caption{Mean labels per decision against the accuracy of the judge, for each setting and $\theta$. PPI, Neyman, Mixture and SLS are shown without refitting. The standard errors of the means are at most 5.5\% of the means.}
\label{fig:sim}
\end{figure}

Without refitting and averaged over the 40 combinations of setting, accuracy and $\theta$, PPI needed 1.16 times the labels of SOLD, Neyman 6.49 times, Mixture 1.64 times, SLS 1.35 times, Uniform 2.45 times and DISCERN 2.88 times (Figure~\ref{fig:sim}, Table~S4). SOLD needed at most 1.09 times the fewest labels in every combination, whereas each of the other methods needed more than 1.5 times the fewest in some combination, from 1.55 times for PPI to 17.15 times for Neyman. When the judge's probabilities are correct, refitting changed the total labels of each design by at most 4\% in setting A and 10\% in setting B (Table~S5).

The advantage of SOLD grew with the accuracy of the judge. Averaged over the eight combinations of setting and $\theta$, PPI needed 1.04 times the labels of SOLD with the 84.1\% accurate judge and 1.29 times with the 99.9\% accurate one, and the other five methods needed between 1.13 and 2.24 times with the 84.1\% accurate judge and between 1.69 and 11.56 times with the 99.9\% accurate one. In setting A SOLD needed between 23 and 159 labels per decision, and in setting B, where every prompt is a candidate, between 128 and 1{,}535, and for every $\theta$ and in both settings it needed fewer labels as the judge became more accurate. SOLD saved the most labels in the hardest case, $\theta=0.07$ in setting B. There it needed 1{,}535 labels with the 84.1\% accurate judge and 518 with the 99.9\% accurate one, and with the 99.9\% accurate judge PPI, Neyman, Mixture and SLS needed between 1.43 and 5.29 times its labels, Uniform 3.60 times and DISCERN 4.18 times. Neyman instead needed more labels as the accuracy of the judge rose from 93.3\% to 99.9\%, from 1{,}426 to 2{,}743 in this case, as Theorem~\ref{thm:threshold} predicts.

\begin{figure}[t]
\centering
\includegraphics[width=\linewidth]{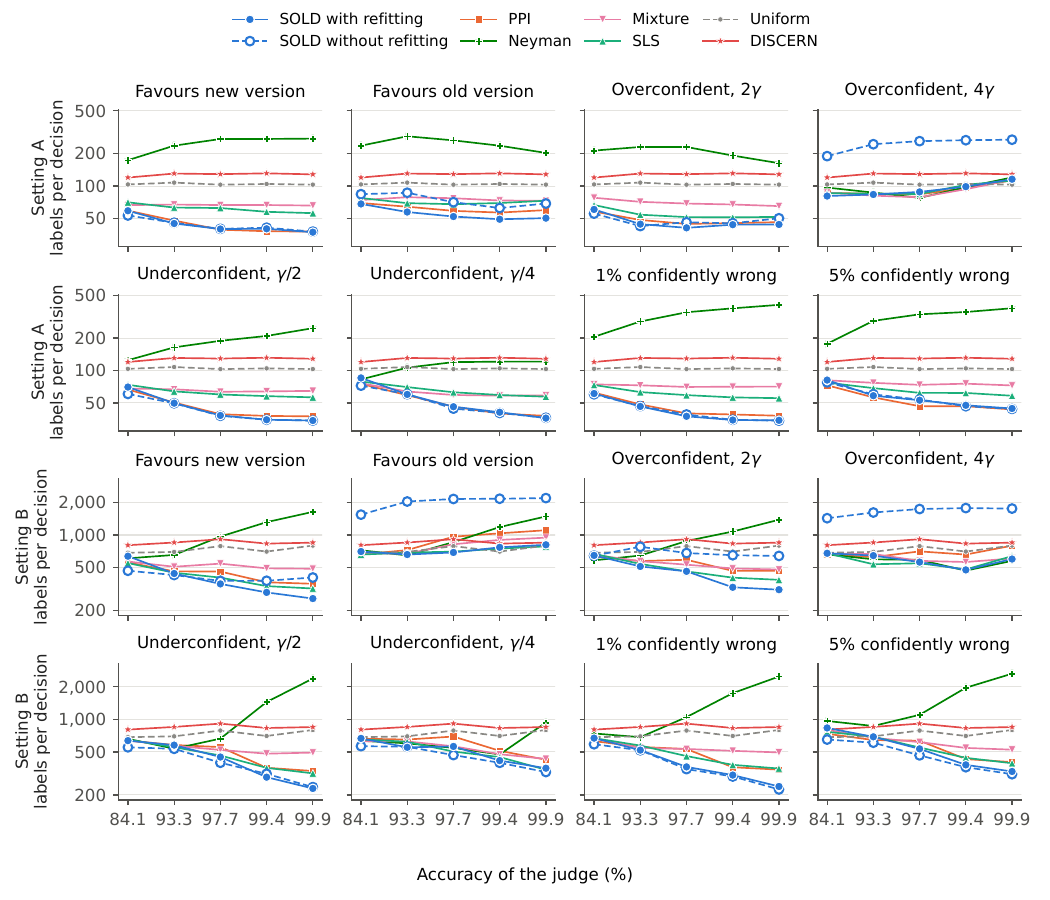}
\caption{Mean labels per decision against the accuracy of the judge when the judge's probabilities are wrong, for each error and setting, with $\theta=0.09$.}
\label{fig:simwrong}
\end{figure}

\subsection{Robustness}\label{sec:simrobust}

Refitting matters when the judge's probabilities are wrong. We kept the pools, the outcomes and the scores of Section~\ref{sec:simdesign}, and hence the accuracy of the judge, and made the judge turn the scores into probabilities wrongly in eight ways. In four of them it applies Bayes' rule as if the scores had mean $c\gamma Y(x)$ instead of $\gamma Y(x)$, with $c=2$ or $4$, which makes the probabilities overconfident, or $c=1/2$ or $1/4$, which makes them underconfident. For the 97.7\% accurate judge, for example, $\gamma=2$, and with $c=4$ it applies Bayes' rule as if the scores had mean $8Y(x)$. In two others it treats every score as $\gamma/2$ higher or lower than it is, half the distance between the mean scores of adjacent outcomes, which makes the probabilities favour the new version, with $Y(x)=1$, or the old one, with $Y(x)=-1$. In the last two it gives 1\% or 5\% of the prompts with $Y(x)\ne0$, chosen at random, probability 0.99 of $-Y(x)$ and 0.005 of each of the other two outcomes. We ran each of the eight errors for the 40 combinations of setting, accuracy and $\theta$. Figure~\ref{fig:simwrong} gives the labels with $\theta=0.09$, and Supplementary Figures~S2, S3 and~S4 give them with $\theta=0$, 0.07 and 0.11. In these figures, PPI, Neyman, Mixture and SLS refit the judge's probabilities.

Compared with the six other methods in Figure~\ref{fig:simwrong}, SOLD with refitting needed the fewest labels in 189 of the 320 combinations of error, setting, accuracy and $\theta$, and was within 5\% of the fewest in 254. With $\theta=0.09$ and the 99.9\% accurate judge, SOLD with or without refitting needed the fewest labels of the methods in Figure~\ref{fig:simwrong} for 12 of the 16 combinations of error and setting. Refitting matters most when the judge's probabilities are overconfident or favour one version. Summed over the five accuracies and the four values of $\theta$ (Tables~S8 and~S9), SOLD without refitting needed 1.95 times the labels of SOLD with refitting in both settings with $c=4$, and 2.36 times in setting B when they favoured the old version. When the probabilities were underconfident or confidently wrong on a few prompts, these totals for SOLD with and without refitting were within 10\% of each other. The largest gap in these totals between SOLD with refitting and another method was in setting B with 5\% of the prompts confidently wrong, where SLS without refitting needed 12\% fewer labels. Only 11 of the 384{,}000 runs made a wrong decision, none of them by SOLD with refitting.

SOLD stayed within 5\% of the fewest labels in 155 of the 160 combinations of setting, accuracy and $\theta$ when we changed the pool size or the tolerance, and needed the fewest in 142 (Table~S7). Its advantage grew with the size of the pool. With $\theta=\delta+0.02$ in setting B and a 99.9\% accurate judge, PPI, Neyman, Mixture and SLS needed between 1.13 and 2.28 times the labels of SOLD for $N=1{,}000$, and between 1.43 and 12.55 times for $N=9{,}000$. Only 2 of the 192{,}000 runs made a wrong decision, none of them by SOLD.

\section{Real data analysis}\label{sec:realdata}\label{sec:evouna}

We replayed the methods on updates in two collections of questions, EVOUNA and MMLU-Pro, in which the outcome of every question is known and is revealed only when a method labels the question. We compare the methods of Section~\ref{sec:sim}. We ran each method 100 times for each question set, update and value of $\delta$, which we call a case. The only randomness is the choice of questions.

\begin{figure}[t]
\centering
\includegraphics[width=\linewidth]{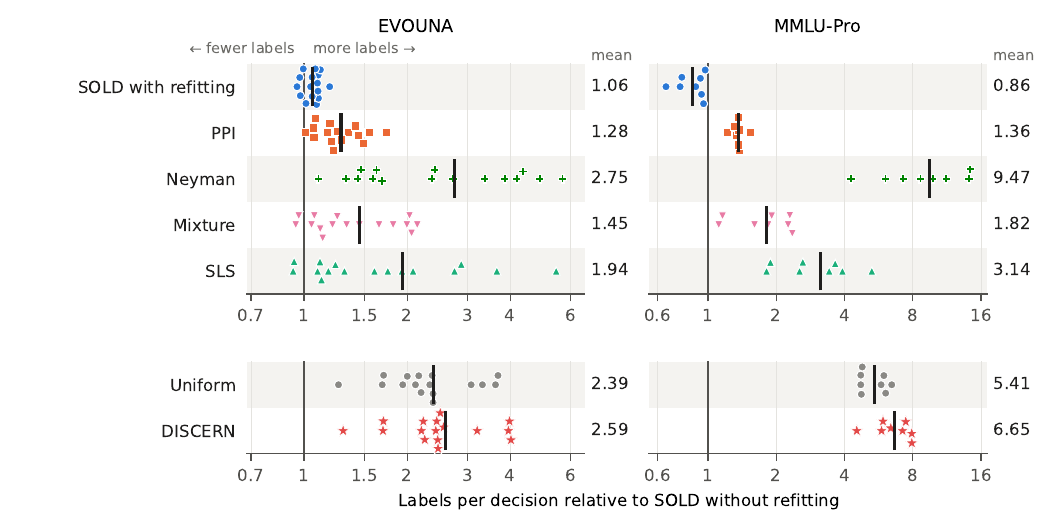}
\caption{Labels per decision on EVOUNA and MMLU-Pro relative to SOLD without refitting. PPI, Neyman, Mixture and SLS are shown without refitting. Each point gives, for one case, the mean labels of the method divided by the mean labels of SOLD without refitting. The bar and the number on the right of each row give the mean of these ratios over the 16 cases of EVOUNA or the 8 cases of MMLU-Pro. The horizontal axes are logarithmic.}
\label{fig:real}
\end{figure}

\subsection{EVOUNA}\label{sec:ev}

We applied SOLD to updates of question-answering systems in EVOUNA \citep{wang2023evaluating}. Each question in EVOUNA comes with reference answers and with the answers of five systems, and human annotators marked the answer of each system as correct or incorrect. The questions come from Natural Questions (NQ) \citep{kwiatkowski2019natural} and TriviaQA (TQ) \citep{joshi2017triviaqa}, and we use 3{,}019 from NQ and 1{,}938 from TQ as two question sets (Supplementary~S6.1). We take four pairs of these systems as updates, the retrieval-based system DPR+FiD to GPT-3.5 (E1), GPT-3.5 to ChatGPT (E2), ChatGPT to GPT-4 (E3) and GPT-4 to Bing Chat (E4).

For an update, a question $x$ to which the two versions give identical answers has $Y(x)=0$ and is not a candidate, which happens for at most 10\% of the questions of a question set (Supplementary~S6.1). For every other question, $Y(x)$ is $1$ if only the answer of the new version is marked correct, $-1$ if only that of the old version is, and $0$ otherwise, and the question is a candidate with $\ell(x)=-1$ and $u(x)=1$. The mean $\theta_g$ of $Y(x)$ over question set $g$ is then the gain in accuracy. We evaluate each update on each of the two question sets separately, and for each question set we declare the update improved, worsened or negligible as in Section~\ref{sec:decisions}, with $\delta=0.05$ or $0.10$ and $\alpha=0.05$. At $\delta=0.05$ the correct decision is improved for E2 and E3 and negligible for E1 and E4, for both question sets, and at $\delta=0.10$ it is negligible for all four updates (Supplementary~S6.1 gives $\theta_g$).

The judge combines two language models, Llama-3.1-8B-Instruct and DeepSeek-V4.1 Flash, which we call J1 and J2. J1 and J2 read a question with its reference answers and one answer, and return the probability $c_v(x)$ that the answer of version $v$ is correct (Supplementary~S6.1). We treat the correctness of the two answers as independent under the judge's predictive distribution, so that for a candidate $x$, $p_+(x)=c_{\rm new}(x)\{1-c_{\rm old}(x)\}$ and $p_-(x)=\{1-c_{\rm new}(x)\}\,c_{\rm old}(x)$ and $m(x)=c_{\rm new}(x)-c_{\rm old}(x)$. Of the probabilities returned by J2, 44\% are exactly 0 or 1, which gives probability zero to an outcome that can occur and leaves the refitting of Supplementary~S1.2 undefined, so we clip all probabilities to $[10^{-6},1-10^{-6}]$, and the $c_v(x)$ of the judge is the average of the clipped probabilities of J1 and J2 on the logit scale. When an answer is called correct if $c_v(x)>1/2$, the judge agreed with the annotators on 95.1\% of the 24{,}785 answers, J1 alone on 90.8\% and J2 alone on 94.9\%, and the area under the ROC curve of the judge was between 0.97 and 0.99 for every update (Table~S12). Supplementary~S6.2 gives the results with J1 or J2 alone as the judge.

SOLD, with or without refitting, needed the fewest labels in 12 of the 16 cases, 10 without refitting and 2 with it, and was within 5\% of the fewest in 13 (Table~S13). Among the methods without refitting, SOLD needed at most 1.08 times the fewest labels in every case, whereas the largest such ratio of the other methods ranged from 1.74 for PPI to 5.69 for Neyman. Averaged over the 16 cases, PPI, Neyman, Mixture and SLS needed between 1.28 and 2.75 times the labels of SOLD, Uniform 2.39 times and DISCERN 2.59 times. Neyman and SLS give labelling probabilities close to zero to the questions for which the judge is nearly certain that $Y(x)=0$, which limits their growth rates whether or not the judge is right (Proposition~\ref{prop:safetybound}). With refitting, SOLD needed on average 1.06 times the labels of SOLD without it, and PPI, Neyman, Mixture and SLS between 1.30 and 1.85 times. Averaged over the 12 cases whose correct decision is negligible, the other methods needed between 1.31 and 3.20 times the labels of SOLD, and over the 4 improved cases between 1.02 and 1.73 times (Supplementary~S6.2). The largest ratio is smaller in the improved cases, where $\theta_g$ exceeds $\delta$ by 0.008 to 0.024, so that labelling every candidate, the most that any method can need, takes on average only 1.94 times the labels of SOLD. No method made a wrong decision in any of the 19{,}200 runs. With J1 or J2 alone as the judge, the other methods needed on average between 1.17 and 2.76 times the labels of SOLD.

\subsection{MMLU-Pro}\label{sec:mmlupro}

MMLU-Pro \citep{wang2024mmlupro} has 12{,}032 multiple-choice questions with a keyed answer each, and its authors released the answers of many language models to every question. We take four updates, Llama-2 from 7B to 13B (M1), Llama-3 to Llama-3.1 at 8B (M2) and at 70B (M3), and Mistral-7B v0.1 to v0.2 (M4), each on the 9{,}797 to 11{,}302 questions on which the released files agree (Supplementary~S6.3). The candidates are the questions on which the two versions choose different options, and $Y(x)$ is defined as in Section~\ref{sec:ev} with the key in place of the annotators, as if an expert checked each answer. We use $\alpha=0.05$ and $\delta=0.02$ or $0.10$. With $\delta=0.02$ the correct decision is improved for every update, and with $\delta=0.10$ it is negligible (Supplementary~S6.3 gives $\theta$).

The judge is Claude Opus 5.5, which sees neither the key nor the answers of the two versions and gives for every option $o$ the probability $q(o)$ that it is the key. For a candidate on which the new and the old version choose $o_{\rm new}$ and $o_{\rm old}$, $p_+(x)=q(o_{\rm new})$ and $p_-(x)=q(o_{\rm old})$ (Supplementary~S6.3). On the questions with $Y(x)\ne0$, $m(x)$ had the sign of $Y(x)$ for 97.5\% to 98.6\% of them, but the probabilities are conservative, and of the answers to which the judge gave a probability between 0.7 and 0.9, 94.3\% were correct.

SOLD with refitting needed the fewest labels in all 8 cases, and SOLD without refitting fewer than every method other than SOLD with refitting (Figure~\ref{fig:real}, Table~S15). Averaged over the cases, PPI needed 1.36 times the labels of SOLD without refitting, Neyman 9.47 times, Mixture 1.82 times, SLS 3.14 times, Uniform 5.41 times and DISCERN 6.65 times, and SOLD with refitting 0.86 times. With refitting, which corrects the conservative probabilities, PPI, Mixture and SLS needed on average 1.47, 2.18 and 3.36 times the labels of SOLD with refitting. Neyman and SLS, which rarely label the questions on which the judge is nearly certain, needed up to 14.3 and 5.3 times the labels of SOLD (Proposition~\ref{prop:safetybound}). SOLD saved more labels here than on EVOUNA, in line with the simulations, in which its advantage grew with the accuracy of the judge. No method made a wrong decision in any of the 9{,}600 runs.

\section{Discussion}\label{sec:disc}
SOLD extends beyond a single update and a fixed judge. When many updates are evaluated, as in the successive rounds of recursive self-improvement, the multiplicity of the tests must be accounted for. Since the wealths of the tests when labelling stops are e-values, the e-Benjamini--Hochberg procedure \citep{wang2022false} controls the false discovery rate among the rejected one-sided hypotheses under arbitrary dependence, and e-LOND \citep{xu2024online} does so when the updates arrive one after another. SOLD also allows the judge to change between draws, as when a self-improving system also updates its evaluator \citep*{chen2026auditing}, since Proposition~\ref{prop:valid} requires only that the predictions $m(x)$ used at step $i$ be $\mathcal F_{i-1}$-measurable.

The lower bound of Theorem~\ref{thm:lb}(a) changes little if a test needs its type I error to be at most $\alpha$ only when the prompts are drawn from its own designs, rather than under every process in $\cN$. When $F(\cdot\mid x)=\mathbf p(x)$ for every $x$, the optimal growth rate of such tests exceeded $G^\star_{\mathbf p}$ by at most 0.74\% in the 48 pools of Supplementary~S3.9, and the two are equal to first order as $\bar\rho\to0$ and in the regime of Proposition~S3 (Supplementary~S4).

Several directions are left for future work. Theorems~\ref{thm:lb} and~\ref{thm:lbD} count draws as $\alpha\to0$, whereas what is paid for is the number of labels at a fixed level $\alpha$, and bounds on this number would sharpen them. When the refitted predictions converge, Theorems~S3 and~S5 give upper bounds relative to their limit, for one test and for the three decisions, and a matching lower bound is a natural next step. \citet{park2025adaptive} average the wealths of tests that give different weights to the judge's predictions. By Proposition~\ref{prop:valid}, each such test is valid under the designs of SOLD, so the two can be combined. If the pool is a sample from a larger population, Theorems~\ref{thm:char} and~\ref{thm:threshold} still compare designs whenever the mean prediction over that population is known, and estimating it is a further extension (Supplementary~S1.3). In recursive self-improvement, successive updates are often evaluated on the same pool. If the response of each version is labelled separately, the labels of the previous version carry over, so a prompt is cheaper to label in later rounds, which a design could take into account.

{\spacingset{1}\footnotesize
\bibliographystyle{chicago-review}
\bibliography{refs}

\begin{thebibliography}{}

\bibitem[\protect\citeauthoryear{Agrawal, Koolen, and Juneja}{Agrawal
  et~al.}{2021}]{agrawal2021optimal}
Agrawal, S., W.~M. Koolen, and S.~Juneja (2021).
\newblock Optimal best-arm identification methods for tail-risk measures.
\newblock In {\em Advances in Neural Information Processing Systems},
  Volume~34, pp.\  25578--25590.

\bibitem[\protect\citeauthoryear{Angelopoulos, Bates, Fannjiang, Jordan, and
  Zrnic}{Angelopoulos et~al.}{2023}]{angelopoulos2023ppi}
Angelopoulos, A.~N., S.~Bates, C.~Fannjiang, M.~I. Jordan, and T.~Zrnic (2023).
\newblock Prediction-powered inference.
\newblock {\em Science\/}~{\em 382\/}(6671), 669--674.

\bibitem[\protect\citeauthoryear{Angelopoulos, Eisenstein, Berant, Agarwal, and
  Fisch}{Angelopoulos et~al.}{2025}]{angelopoulos2025cost}
Angelopoulos, A.~N., J.~Eisenstein, J.~Berant, A.~Agarwal, and A.~Fisch (2025).
\newblock Cost-optimal active {AI} model evaluation.
\newblock arXiv:2506.07949.

\bibitem[\protect\citeauthoryear{Ao, Chen, Gao, Li, and Simchi-Levi}{Ao
  et~al.}{2026}]{ao2026best}
Ao, R., H.~Chen, S.~Gao, H.~Li, and D.~Simchi-Levi (2026).
\newblock Designing service systems from textual evidence.
\newblock arXiv:2601.21471.

\bibitem[\protect\citeauthoryear{Balachandran}{Balachandran}{2026}]{balachandran2026pay}
Balachandran, V.~B. (2026).
\newblock Pay only for disagreement: Certified no-regression verdicts for model
  updates with matching label-complexity bounds.
\newblock arXiv:2609.17560.

\bibitem[\protect\citeauthoryear{Breiman}{Breiman}{1961}]{breiman1961}
Breiman, L. (1961).
\newblock Optimal gambling systems for favorable games.
\newblock In {\em Proceedings of the Fourth Berkeley Symposium on Mathematical
  Statistics and Probability}, Volume~1, pp.\  65--78. University of California
  Press.

\bibitem[\protect\citeauthoryear{Chen, Wang, and Qu}{Chen
  et~al.}{2026}]{chen2026recursive}
Chen, M., L.~Wang, and B.~Qu (2026).
\newblock Recursive self-improvement in {AI}: From bounded self-refinement to
  autonomous research loops.
\newblock arXiv:2607.07663.

\bibitem[\protect\citeauthoryear{Chen, Zhu, and Zheng}{Chen
  et~al.}{2026}]{chen2026auditing}
Chen, Z., D.~Zhu, and L.~N. Zheng (2026).
\newblock Auditing the product, not the recursion: The evaluation evidence
  behind recursive self-improvement claims.
\newblock arXiv:2610.27610.

\bibitem[\protect\citeauthoryear{Chernoff}{Chernoff}{1959}]{chernoff1959sequential}
Chernoff, H. (1959).
\newblock Sequential design of experiments.
\newblock {\em Annals of Mathematical Statistics\/}~{\em 30\/}(3), 755--770.

\bibitem[\protect\citeauthoryear{Cover}{Cover}{1991}]{cover1991universal}
Cover, T.~M. (1991).
\newblock Universal portfolios.
\newblock {\em Mathematical Finance\/}~{\em 1\/}(1), 1--29.

\bibitem[\protect\citeauthoryear{Csillag, Struchiner, and Goedert}{Csillag
  et~al.}{2025}]{csillag2025prediction}
Csillag, D., C.~J. Struchiner, and G.~T. Goedert (2025).
\newblock Prediction-powered e-values.
\newblock In {\em Proceedings of the 42nd International Conference on Machine
  Learning}, Volume 267 of {\em Proceedings of Machine Learning Research}.

\bibitem[\protect\citeauthoryear{Dorner, Nastl, and Hardt}{Dorner
  et~al.}{2025}]{dorner2025limits}
Dorner, F.~E., V.~Y. Nastl, and M.~Hardt (2025).
\newblock Limits to scalable evaluation at the frontier: {LLM} as judge won't
  beat twice the data.
\newblock In {\em International Conference on Learning Representations}.

\bibitem[\protect\citeauthoryear{Gao, Hu, Liu, and Zhou}{Gao
  et~al.}{2026}]{gao2026empirical}
Gao, H., J.~Hu, P.~Liu, and W.~Zhou (2026).
\newblock Anytime-valid empirical likelihood inference from {LLM} annotations
  with adaptive human auditing.
\newblock arXiv:2611.15197.

\bibitem[\protect\citeauthoryear{Garivier and Kaufmann}{Garivier and
  Kaufmann}{2016}]{garivier2016optimal}
Garivier, A. and E.~Kaufmann (2016).
\newblock Optimal best arm identification with fixed confidence.
\newblock In {\em Proceedings of the 29th Conference on Learning Theory},
  Volume~49 of {\em Proceedings of Machine Learning Research}, pp.\  998--1027.

\bibitem[\protect\citeauthoryear{Ham, Zhao, Jasin, and Yang}{Ham
  et~al.}{2026}]{ham2026human}
Ham, D.~W., X.~Zhao, S.~Jasin, and F.~Yang (2026).
\newblock Human-{AI}-powered hypothesis testing: Cost-aware selective {AI}
  scoring and sequential human escalation.
\newblock arXiv:2609.28859.

\bibitem[\protect\citeauthoryear{Honda and Takemura}{Honda and
  Takemura}{2010}]{honda2010asymptotically}
Honda, J. and A.~Takemura (2010).
\newblock An asymptotically optimal bandit algorithm for bounded support
  models.
\newblock In {\em Proceedings of the 23rd Conference on Learning Theory}, pp.\
  67--79.

\bibitem[\protect\citeauthoryear{Hsu and Shekhar}{Hsu and
  Shekhar}{2026}]{hsu2026efficient}
Hsu, C.-Y. and S.~Shekhar (2026).
\newblock Efficient sequential evaluation of large language models.
\newblock arXiv:2607.17409.

\bibitem[\protect\citeauthoryear{Joshi, Choi, Weld, and Zettlemoyer}{Joshi
  et~al.}{2017}]{joshi2017triviaqa}
Joshi, M., E.~Choi, D.~Weld, and L.~Zettlemoyer (2017).
\newblock {TriviaQA}: A large scale distantly supervised challenge dataset for
  reading comprehension.
\newblock In {\em Proceedings of the 55th Annual Meeting of the Association for
  Computational Linguistics (Volume 1: Long Papers)}, pp.\  1601--1611.
  Association for Computational Linguistics.

\bibitem[\protect\citeauthoryear{Kaufmann, Capp{\'e}, and Garivier}{Kaufmann
  et~al.}{2016}]{kaufmann2016complexity}
Kaufmann, E., O.~Capp{\'e}, and A.~Garivier (2016).
\newblock On the complexity of best-arm identification in multi-armed bandit
  models.
\newblock {\em Journal of Machine Learning Research\/}~{\em 17\/}(1), 1--42.

\bibitem[\protect\citeauthoryear{Kelly}{Kelly}{1956}]{kelly1956}
Kelly, Jr., J.~L. (1956).
\newblock A new interpretation of information rate.
\newblock {\em Bell System Technical Journal\/}~{\em 35\/}(4), 917--926.

\bibitem[\protect\citeauthoryear{Kwiatkowski, Palomaki, Redfield, Collins,
  Parikh, Alberti, Epstein, Polosukhin, Devlin, Lee, Toutanova, Jones, Kelcey,
  Chang, Dai, Uszkoreit, Le, and Petrov}{Kwiatkowski
  et~al.}{2019}]{kwiatkowski2019natural}
Kwiatkowski, T., J.~Palomaki, O.~Redfield, M.~Collins, A.~Parikh, C.~Alberti,
  D.~Epstein, I.~Polosukhin, J.~Devlin, K.~Lee, K.~Toutanova, L.~Jones,
  M.~Kelcey, M.-W. Chang, A.~M. Dai, J.~Uszkoreit, Q.~Le, and S.~Petrov (2019).
\newblock Natural {Q}uestions: A benchmark for question answering research.
\newblock {\em Transactions of the Association for Computational
  Linguistics\/}~{\em 7}, 453--466.

\bibitem[\protect\citeauthoryear{Lin, Krever, {\v{C}}ern{\'y}, and Xu}{Lin
  et~al.}{2026}]{lin2026not}
Lin, F., N.~Krever, J.~{\v{C}}ern{\'y}, and L.~Xu (2026).
\newblock Not all label errors are equal: Efficient task-aware auditing with
  confidence sequences.
\newblock In {\em Advances in Neural Information Processing Systems},
  Volume~39.

\bibitem[\protect\citeauthoryear{Mondal, Banik, Apurba, and Mahmud}{Mondal
  et~al.}{2026}]{mondal2026old}
Mondal, J.~J., M.~Banik, M.~S.~A. Apurba, and M.~M.~A. Mahmud (2026).
\newblock When can old evaluations certify a new model? {L}abel-efficient
  release decisions under evaluator drift.
\newblock arXiv:2609.32267.

\bibitem[\protect\citeauthoryear{Naghshvar and Javidi}{Naghshvar and
  Javidi}{2013}]{naghshvar2013active}
Naghshvar, M. and T.~Javidi (2013).
\newblock Active sequential hypothesis testing.
\newblock {\em Annals of Statistics\/}~{\em 41\/}(6), 2703--2738.

\bibitem[\protect\citeauthoryear{Neyman}{Neyman}{1934}]{neyman1934}
Neyman, J. (1934).
\newblock On the two different aspects of the representative method: The method
  of stratified sampling and the method of purposive selection.
\newblock {\em Journal of the Royal Statistical Society\/}~{\em 97\/}(4),
  558--625.

\bibitem[\protect\citeauthoryear{Park, Zecchin, and Simeone}{Park
  et~al.}{2025}]{park2025adaptive}
Park, S., M.~Zecchin, and O.~Simeone (2025).
\newblock Adaptive prediction-powered {AutoEval} with reliability and
  efficiency guarantees.
\newblock In {\em Advances in Neural Information Processing Systems}.

\bibitem[\protect\citeauthoryear{Ramdas, Gr{\"u}nwald, Vovk, and Shafer}{Ramdas
  et~al.}{2023}]{ramdas2023game}
Ramdas, A., P.~Gr{\"u}nwald, V.~Vovk, and G.~Shafer (2023).
\newblock Game-theoretic statistics and safe anytime-valid inference.
\newblock {\em Statistical Science\/}~{\em 38\/}(4), 576--601.

\bibitem[\protect\citeauthoryear{Sandoval, Waudby-Smith, and Jordan}{Sandoval
  et~al.}{2026}]{sandoval2026multi}
Sandoval, R.~J., I.~Waudby-Smith, and M.~I. Jordan (2026).
\newblock Multi-armed sequential hypothesis testing by betting.
\newblock arXiv:2603.17925.

\bibitem[\protect\citeauthoryear{Sawade, Landwehr, and Scheffer}{Sawade
  et~al.}{2012}]{sawade2012active}
Sawade, C., N.~Landwehr, and T.~Scheffer (2012).
\newblock Active comparison of prediction models.
\newblock In {\em Advances in Neural Information Processing Systems},
  Volume~25.

\bibitem[\protect\citeauthoryear{Schuirmann}{Schuirmann}{1987}]{schuirmann1987comparison}
Schuirmann, D.~J. (1987).
\newblock A comparison of the two one-sided tests procedure and the power
  approach for assessing the equivalence of average bioavailability.
\newblock {\em Journal of Pharmacokinetics and Biopharmaceutics\/}~{\em
  15\/}(6), 657--680.

\bibitem[\protect\citeauthoryear{Settles}{Settles}{2009}]{settles2009active}
Settles, B. (2009).
\newblock Active learning literature survey.
\newblock Computer Sciences Technical Report 1648, University of
  Wisconsin--Madison.

\bibitem[\protect\citeauthoryear{Shafer}{Shafer}{2021}]{shafer2021testing}
Shafer, G. (2021).
\newblock Testing by betting: A strategy for statistical and scientific
  communication.
\newblock {\em Journal of the Royal Statistical Society, Series A\/}~{\em
  184\/}(2), 407--431.

\bibitem[\protect\citeauthoryear{Shekhar}{Shekhar}{2025}]{shekhar2025optimal}
Shekhar, S. (2025).
\newblock Optimal anytime-valid tests for composite nulls.
\newblock arXiv:2512.20039.

\bibitem[\protect\citeauthoryear{Spertus, Sridhar, and Stark}{Spertus
  et~al.}{2024}]{spertus2024sequential}
Spertus, J.~V., M.~Sridhar, and P.~B. Stark (2024).
\newblock Sequential stratified inference for the mean.
\newblock arXiv:2409.06680.

\bibitem[\protect\citeauthoryear{Ville}{Ville}{1939}]{ville1939}
Ville, J. (1939).
\newblock {\em {\'E}tude critique de la notion de collectif}.
\newblock Paris: Gauthier-Villars.

\bibitem[\protect\citeauthoryear{Wang, Cheng, Guo, Yue, Ding, Xu, Wang, Hu,
  Zhang, and Zhang}{Wang et~al.}{2023}]{wang2023evaluating}
Wang, C., S.~Cheng, Q.~Guo, Y.~Yue, B.~Ding, Z.~Xu, Y.~Wang, X.~Hu, Z.~Zhang,
  and Y.~Zhang (2023).
\newblock Evaluating open-{QA} evaluation.
\newblock In {\em Advances in Neural Information Processing Systems (Datasets
  and Benchmarks Track)}, Volume~36, pp.\  77013--77042.

\bibitem[\protect\citeauthoryear{Wang and Ramdas}{Wang and
  Ramdas}{2022}]{wang2022false}
Wang, R. and A.~Ramdas (2022).
\newblock False discovery rate control with e-values.
\newblock {\em Journal of the Royal Statistical Society, Series B\/}~{\em
  84\/}(3), 822--852.

\bibitem[\protect\citeauthoryear{Wang, Ma, Zhang, Ni, Chandra, et~al.}{Wang
  et~al.}{2024}]{wang2024mmlupro}
Wang, Y., X.~Ma, G.~Zhang, Y.~Ni, A.~Chandra, et~al. (2024).
\newblock {MMLU-Pro}: A more robust and challenging multi-task language
  understanding benchmark.
\newblock In {\em Advances in Neural Information Processing Systems (Datasets
  and Benchmarks Track)}.

\bibitem[\protect\citeauthoryear{Waudby-Smith and Ramdas}{Waudby-Smith and
  Ramdas}{2024}]{waudbysmith2024estimating}
Waudby-Smith, I. and A.~Ramdas (2024).
\newblock Estimating means of bounded random variables by betting.
\newblock {\em Journal of the Royal Statistical Society, Series B\/}~{\em
  86\/}(1), 1--27.

\bibitem[\protect\citeauthoryear{Wu, Nair, and Cand{\`e}s}{Wu
  et~al.}{2026}]{wu2026efficient}
Wu, S., Y.~Nair, and E.~J. Cand{\`e}s (2026).
\newblock Efficient evaluation of {LLM} performance with statistical
  guarantees.
\newblock arXiv:2601.20251.

\bibitem[\protect\citeauthoryear{Xu, Karampatziakis, and Mineiro}{Xu
  et~al.}{2024}]{xu2024active}
Xu, Z., N.~Karampatziakis, and P.~Mineiro (2024).
\newblock Active, anytime-valid risk controlling prediction sets.
\newblock In {\em Advances in Neural Information Processing Systems},
  Volume~37, pp.\  60110--60132.

\bibitem[\protect\citeauthoryear{Xu and Ramdas}{Xu and
  Ramdas}{2024}]{xu2024online}
Xu, Z. and A.~Ramdas (2024).
\newblock Online multiple testing with e-values.
\newblock In {\em Proceedings of the 27th International Conference on
  Artificial Intelligence and Statistics}, Volume 238 of {\em Proceedings of
  Machine Learning Research}, pp.\  3997--4005.

\bibitem[\protect\citeauthoryear{Zhou, Ye, Chen, Li, and Liu}{Zhou
  et~al.}{2026}]{li2026celeus}
Zhou, Z., Z.~Ye, Z.~Chen, B.~Li, and F.~Liu (2026).
\newblock {CELEUS}: Certifiable and efficient {LLM} evaluation via e-processes.
\newblock arXiv:2606.20820.

\bibitem[\protect\citeauthoryear{Zrnic and Cand{\`e}s}{Zrnic and
  Cand{\`e}s}{2024}]{zrnic2024active}
Zrnic, T. and E.~J. Cand{\`e}s (2024).
\newblock Active statistical inference.
\newblock In {\em Proceedings of the 41st International Conference on Machine
  Learning}, Volume 235 of {\em Proceedings of Machine Learning Research}, pp.\
   62993--63010.

\end{thebibliography}
}

\end{document}